\documentclass{article}
\PassOptionsToPackage{numbers, compress}{natbib}

\usepackage[utf8]{inputenc}
\usepackage[T1]{fontenc}
\usepackage{url}
\usepackage{booktabs}
\usepackage{graphicx}
\usepackage{amsmath}
\usepackage{amssymb}
\usepackage{nicefrac}
\usepackage{microtype}
\usepackage{xcolor}

\usepackage{graphicx} 
\usepackage{amsmath}
\usepackage{amssymb}
\usepackage{amsfonts}
\usepackage{booktabs}      
\usepackage[table]{xcolor} 
\usepackage{colortbl}      
\usepackage{subcaption}
\usepackage{natbib}
\usepackage{url}

 \usepackage[preprint]{neurips_2026}

\definecolor{lightgray}{gray}{0.92}
\definecolor{lightcyan}{rgb}{0.88, 1.0, 1.0}
\usepackage{multirow}
\usepackage{pifont} 
\usepackage{xcolor}
\definecolor{vividgreen}{HTML}{00CC00}
\newcommand{\cmark}{\ding{51}}
\newcommand{\xmark}{\ding{55}}

\usepackage{hyperref}

\title{Does the Question Really Matter? Training-Free Data Selection for Vision-Language SFT}

\author{%
\begin{tabular}{c}
\textbf{Peng Sun}$^{1}$ \quad
\textbf{Yi Yang}$^{1}$ \quad
\textbf{Huawen Shen}$^{2}$ \\
\textbf{Yi Ban}$^{1}$ \quad
\textbf{Yanbo Wang}$^{3}$ \quad
\textbf{Yuqiang Li}$^{4}$ \quad
\textbf{Tianfan Fu}$^{1}$\thanks{Corresponding author.}
\end{tabular}
\\[0.8em]
\small
$^{1}$Nanjing University \quad
$^{2}$Institute of Information Engineering
\\
\small
$^{3}$North University of China \quad
$^{4}$Shanghai Artificial Intelligence Laboratory
}

\begin{document}

\maketitle
\begin{abstract}
Visual instruction tuning is crucial for improving vision-language large models (VLLMs). However, many samples can be solved via linguistic patterns or common-sense shortcuts, without genuine cross-modal reasoning, limiting the effectiveness of multimodal learning. Prior data selection methods often rely on costly proxy model training and focus on difficulty or diversity, failing to capture a sample’s true contribution to vision-language joint reasoning. In this paper, we propose CVS, a training-free data selection method based on the insight that, for high-quality multimodal samples, introducing the question should substantially alter the model’s assessment of answer validity given an image. CVS leverages a frozen VLLM as an evaluator and measures the discrepancy in answer validity with and without conditioning on the question, enabling the identification of samples that require vision-language joint reasoning while filtering semantic-conflict noise. Experiments on Vision-Flan and The Cauldron show that CVS achieves solid performance across datasets. On Vision-Flan, CVS outperforms full-data training by 3.5\% and 4.8\% using only 10\% and 15\% of the data, respectively, and remains robust on the highly heterogeneous Cauldron dataset. Moreover, CVS reduces computational cost by 17.3\% and 44.4\% compared to COINCIDE and XMAS. The code will be made publicly available upon acceptance of the paper.
\end{abstract}

\section{Introduction}
One of the key sources of the capabilities of vision-language large models (VLLMs) is Visual Instruction Tuning (VIT)~\cite{liu2023visual}, which enables models to better align natural language instructions with visual content across a wide range of tasks~\cite{li2023visionlanguageinstructiontuningreview,zhou2025learninginstructvisualinstruction}. However, a gradually emerging yet underexplored issue is that, during visual instruction tuning, not all samples that are formally multimodal actually require genuine cross-modal reasoning~\cite{wu2025mitigatingmodalimbalancemultimodal}.

Among existing datasets, many samples can be answered correctly with little or no reliance on visual content, as models exploit linguistic patterns or common-sense priors~\cite{lu2025eliminating,sim2025can}. Such samples provide weak cross-modal supervision and encourage models to adopt linguistic shortcuts rather than visual reasoning, ultimately degrading sensitivity to visual evidence~\cite{yuksekgonul2023visionlanguagemodelsbehavelike}. This issue is exacerbated in real-world data construction pipelines, where heterogeneous collection methods introduce supervision noise and image–text misalignment, rendering simple data scaling insufficient for improving vision-language models~\cite{chen2023weaklysupervisedvisionandlanguagepretraining,yan2025data}.

Existing data selection methods can be categorized into score-based and clustering-based methods. The former typically assign scalar scores to samples based on criteria such as difficulty, gradient contribution, or similarity, but often treat sample utility as an independently evaluable property, making it difficult to distinguish samples that genuinely require vision-language joint reasoning from those that can be solved via linguistic shortcuts~\cite{lee2025vlind}. The latter aim to improve data diversity through clustering or deduplication; however, diversity alone does not guarantee that a question meaningfully constrains its answer~\cite{wang2024diversitymeasurementsubsetselection}. Moreover, most existing methods rely on additional proxy models or complex post-processing pipelines, which introduce substantial computational overhead in large-scale data settings~\cite{albalak2024surveydataselectionlanguage}.

We posit that a sample's value implies Visual Necessity: the question must provide non-redundant information gain regarding the answer's validity, conditioned on the visual context. Only when the question substantively affects the decision process and forces the model to jointly reason over visual content and linguistic instructions does the supervision signal become effective. This conditional dependence captures the essence of cross-modal supervision, yet has not been explicitly modeled in existing data selection methods.

Recent advances in VLLMs show that modern models are already capable of making stable zero-shot judgments about answer validity given an image and text~\cite{danish2025comprehensive,nagar2024zero}. Motivated by this observation, we adopt a reverse perspective. Instead of training an additional proxy model to assess sample quality, we directly leverage a frozen VLLM as an evaluator and use changes in its judgment behavior to characterize the supervisory value of a sample. Based on this insight, we propose Conditional Verdict Shift (CVS), a training-free data selection method. CVS evaluates sample value by comparing the model’s judgments of answer validity under two conditions. One condition uses the full context, including the image, question, and answer, while the other removes the question and retains only the image and the answer. If the inclusion of the question increases the model’s acceptance of the answer without strengthening its tendency to reject it, the sample is regarded as semantically consistent and provides effective cross-modal constraints. Otherwise, it is more likely to reflect semantic conflicts or noisy supervision. Through this conditional verdict shift, CVS filters low-quality samples and prioritizes high-value ones that require genuine cross-modal reasoning, without performing any additional training.

Our main contributions are summarized as: 
\begin{itemize}
\item We identify a critical yet often overlooked issue in visual instruction data: a large number of samples do not genuinely rely on vision-language joint reasoning, but can instead be solved via linguistic shortcuts, thereby weakening the effectiveness of cross-modal learning.
\item We propose Conditional Verdict Shift (CVS), a training-free data selection method based on a novel perspective that models the effectiveness of VIT data through the conditional influence of the question on answer validity. This allows efficient identification of high-value samples requiring genuine joint reasoning.
\item We systematically evaluate CVS on two widely recognized VIT datasets, demonstrating its advantages in performance, stability, and computational efficiency. On VISION-FLAN, using only 10\% and 15\% of samples selected by CVS yields 3.5\% and 4.8\% higher performance than full-data training, respectively. Furthermore, on The Cauldron, CVS reduces computational costs by 17.3\% and 44.4\% compared to COINCIDE and XMAS.
\end{itemize}

\section{Method}




\subsection{Problem Formulation}
\label{sec:problem_formulation}

Appendix~\ref{sec:preliminaries} analyzes data selection from a distributional perspective and shows that, when a sufficiently strong evaluator is available, satisfactory generalization can be achieved using only a small selected subset. This section does not introduce additional theoretical objectives or optimization assumptions. Rather, it reformulates the data selection problem from an operational discriminative perspective, thereby clarifying the core question that motivates the subsequent CVS design.

We consider a candidate training pool
\begin{equation}
\mathcal{S}_{\mathrm{pool}}=\{s_i\}_{i=1}^{N}, 
\quad s_i=(I_i,Q_i,A_i),
\end{equation}
where $I_i$, $Q_i$, and $A_i$ denote the image, question, and answer of the $i$-th sample, respectively. Since candidate samples may contain heterogeneous supervision signals, such as linguistic shortcuts, weak visual grounding, or semantic conflicts, we do not explicitly model the target data distribution. Instead, we introduce a binary selection variable
\begin{equation}
m_i \in \{0,1\},
\end{equation}
where $m_i=1$ indicates that $s_i$ is retained for training. The selected training subset is therefore
\begin{equation}
\mathcal{S}_{\mathrm{train}}
=
\{s_i \in \mathcal{S}_{\mathrm{pool}} \mid m_i=1\}.
\end{equation}

To characterize the desired selection criterion, we define an abstract sample value function $\mathcal{V}(s_i)$, which measures the extent to which a sample contributes useful vision-language learning signals. Under a budget constraint $K$, data selection can be conceptually written as
\begin{equation}
\max_{\{m_i\}_{i=1}^{N}} \sum_{i=1}^{N} m_i \mathcal{V}(s_i),
\quad \text{s.t. } \sum_{i=1}^{N} m_i \le K,\; m_i \in \{0,1\}.
\end{equation}
This formulation is not intended as an optimization objective to be solved directly. Rather, it specifies the type of sample value that an effective selection method should approximate. In this work, the key question is how to instantiate $\mathcal{V}(\cdot)$ in a computable and training-free manner, while keeping it closely aligned with vision-language joint reasoning. The proposed Conditional Verdict Shift (CVS), introduced in Section~\ref{sec:conditional_verdict_shift}, addresses this question by using a frozen vision-language model to measure discriminative changes under different conditioning contexts.

\begin{figure*}[t]
    \centering

    \begin{subfigure}[t]{0.75\textwidth}
        \centering
        \includegraphics[width=\linewidth]{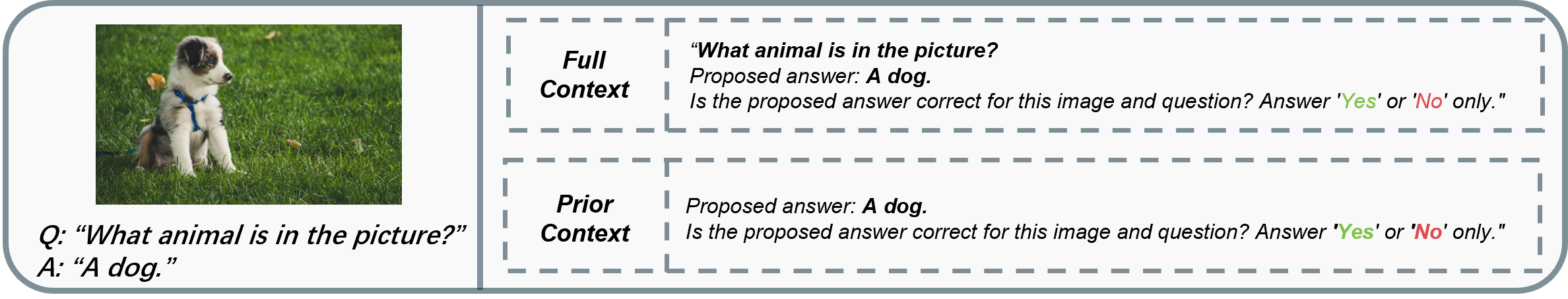}
        \caption{Prompt construction.}
        \label{fig:method:prompt}
    \end{subfigure}

    \vspace{0.5em}

    \begin{subfigure}[t]{0.75\textwidth}
        \centering
        \includegraphics[width=\linewidth]{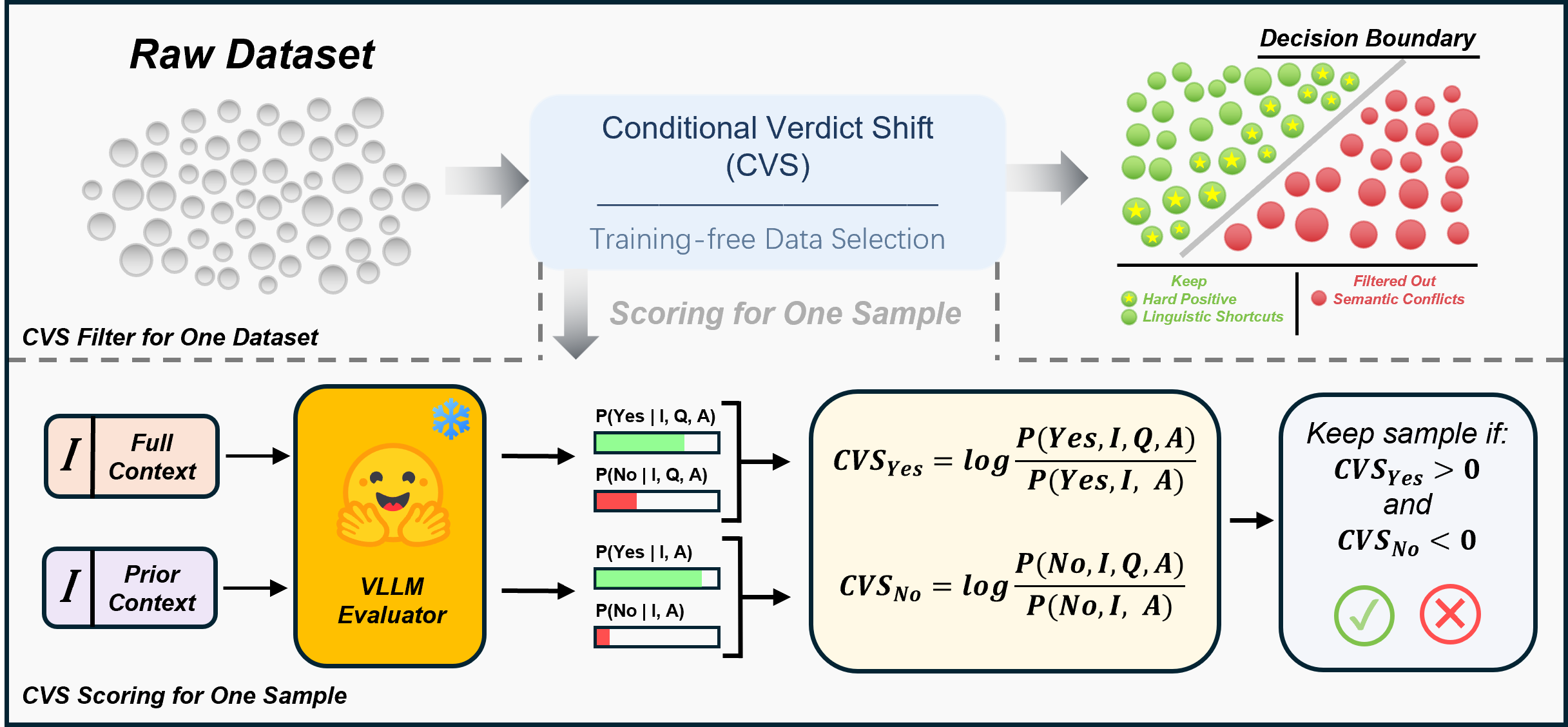}
        \caption{CVS pipeline.}
        \label{fig:method:overview}
    \end{subfigure}

    \caption{Overview of CVS. (a) Prompt construction. (b) CVS uses a frozen vision-language model to estimate conditional verdict shifts, filter noisy samples, and prioritize informative samples near the decision boundary.}
    \label{fig:method}
\end{figure*}

\subsection{Conditional Verdict Shift (CVS)}
\label{sec:conditional_verdict_shift}

In our implementation, we use a frozen, well-trained VLLM as an intrinsic evaluator that evaluates the validity of an answer $A$ given its context. This is formulated as a single-step binary classification task over $\mathcal{Y} = \{\textsc{Yes}, \textsc{No}\}$. Rather than relying on absolute likelihoods, we characterize the extent to which the textual query $Q$ modulates the model's judgment regarding the validity of the answer $A$. This design draws inspiration from established methodology~\cite{Ethayarajh2021UnderstandingDD} that uses conditional V-information to evaluate the semantic quality of instances via pointwise mutual information estimates.

The overall CVS prompt construction and pipeline are illustrated in Figures~\ref{fig:method:prompt} and~\ref{fig:method:overview}, respectively, with detailed prompt templates provided in Appendix~\ref{Templates}. For binary predictions, we directly use the model’s output probabilities for $\{\textsc{Yes}, \textsc{No}\}$ without additional softmax normalization. Since different answer-token choices and prompt templates may affect the resulting model-assigned probabilities, we examine the sensitivity of CVS to these design choices in Appendix~\ref{app:prompt_sensitivity} to assess its robustness.

\paragraph{Conditional Affirmation Shift}

We measure the modulation of confidence induced by the query $Q$. Specifically, we compute the probability of the token `\textsc{Yes}' under two conditions: the full context $(I,Q,A)$ and a reduced context where the query $Q$ is removed. We formalize this comparison as the \textit{Conditional Affirmation Shift} ($\text{CVS}_{\textsc{Yes}}$), defined as the logarithmic ratio:

\begin{equation}
    \text{CVS}_{\textsc{Yes}} = \log \frac{P(\textsc{Yes} \mid I, Q, A)}{P(\textsc{Yes} \mid I, A)}.
\end{equation}
This metric quantifies the information gain provided by the query $Q$ regarding the validity of the answer. A positive shift, i.e., $\text{CVS}_{\textsc{Yes}} > 0$, indicates that the inclusion of the question reinforces the model's belief in the answer's correctness, thereby signifying a strong semantic alignment between the query $Q$ and the response $A$.

\paragraph{Conditional Rejection Shift}
\label{sec:conditional_rejection_shift}

To characterize the ``inverse impact'' of the textual query on the answer's validity, we further investigate the relative variation in the model's rejection confidence. Analogously to the affirmation shift, we define the \textit{Conditional Rejection Shift} ($\text{CVS}_{\textsc{No}}$) as:
\begin{equation}
    \text{CVS}_{\textsc{No}} = \log \frac{P(\textsc{No} \mid I, Q, A)}{P(\textsc{No} \mid I, A)}.
\end{equation}
This metric measures whether the introduction of the query $Q$ reinforces the model's tendency to classify the answer as ``invalid'' or ``incorrect''. A positive shift ($\text{CVS}_{\textsc{No}} > 0$) implies that the query context elevates the probability of rejection, signaling a semantic conflict or misalignment between the question and the answer (e.g., hallucinations or irrelevant responses).


\paragraph{Filtering Protocol}
\label{sec:filtering_protocol}

Combining gains from both affirmation and rejection, we establish a principled data filtering protocol. Specifically, we construct an admissible dataset $\mathcal{D}_{\text{select}}$ by retaining samples that satisfy the following alignment constraints:
\begin{equation}
    \mathcal{C}_{\text{align}}: \quad (\text{CVS}_{\textsc{Yes}} > 0) \land (\text{CVS}_{\textsc{No}} < 0).
\end{equation}
This protocol enforces semantic consistency by ensuring answer validity while suppressing rejection, filtering inconsistent or misleading QA pairs. 
We keep the affirmation and rejection shifts as two separate constraints because they capture answer support and semantic conflict from different directions. To examine whether this separation is necessary in practice, we further compare variants using only $\text{CVS}_{\textsc{Yes}}$, only $\text{CVS}_{\textsc{No}}$, and three coupled log-odds variants, with results reported in Appendix~\ref{app:yes_no_complementarity}.

For $\text{CVS}_{\textsc{Yes}}$ and $\text{CVS}_{\textsc{No}}$, instead of using more specialized thresholding schemes, we adopt zero as a natural and effective threshold, since it directly separates positive from negative shifts. With this choice, we retain $59.29\%$ and $63.97\%$ of samples on the two datasets used in subsequent experiments, respectively.


\paragraph{Preference for Hard Positive Samples}
\label{subsec:pre_for_hps}

Building on our filter protocol, and contrary to intuition, we prioritize samples with lower positive $\text{CVS}_{\textsc{Yes}}$ rather than maximizing this metric. A high $\text{CVS}_{\textsc{Yes}}$ suggests that the model can resolve the reasoning path $Q \to I \to A$ with little effort, often due to strong linguistic correlations or commonsense shortcuts that allow it to ignore the visual input $I$ and rely on language priors alone. In contrast, we focus on instances that require joint reasoning, where $I$ provides essential contextual grounding and $Q$ specifies the directive information. A lower $\text{CVS}_{\textsc{Yes}}$ indicates that the model validates the answer without excessive confidence, reflecting a non-trivial inference process. These samples lie closer to the decision boundary and, during training, force explicit integration of visual features and textual semantics, yielding substantially stronger gradient learning signals than high-confidence samples.

\section{Experiments}
\label{sec:experiments}

In this section, we describe the experimental setup in Section~\ref{sec:setup} and analyze the main results in Section~\ref{sec:main_results}.

\subsection{Experimental Setup}
\label{sec:setup}

\paragraph{Visual Instruction Tuning Datasets}
We conduct experiments on two widely used ViT datasets, Vision-Flan~\cite{xu2024visionflanscalinghumanlabeledtasks} and The Cauldron~\cite{laurenccon2024matters} (see Appendix~\ref{Dataset_dtl} for details).

\paragraph{Models and Training Details}
\label{sec:models_training}
We use the pre-trained LLaVA-1.5-7B model~\cite{liu2024improved} as the target model and Qwen2.5-VL-7B-Instruct model~\cite{bai2025qwen25vltechnicalreport} as the evaluator, with inference accelerated by vLLM~\cite{kwon2023efficient}. Unless otherwise specified (Section~\ref{subsec:target_model}), we follow the official LLaVA-1.5 LoRA fine-tuning hyperparameters. All experiments are conducted on 4 NVIDIA RTX A6000 GPUs.
\paragraph{Baselines}
\label{sec:baselines}

In the main experiments, we compare CVS with CLIP-Score~\cite{hessel2021clipscore}, EL2N~\cite{paul2021deep}, SemDeDup~\cite{abbas2023semdedupdataefficientlearningwebscale}, D2 Pruning~\cite{maharana2023d2pruningmessagepassing}, Random, COINCIDE~\cite{lee2024coincide}, and XMAS~\cite{naharas2025dataselectionfinetuningvision}. Notably, COINCIDE (2024) and XMAS (2025) achieved state-of-the-art performance at the time of their introduction. Detailed hyperparameter settings for these baselines are provided in Appendix~\ref{Baseline_hyperparam}. We also include two recent data selection methods, SCALSELECT~\cite{wu2026scalselect} and PRISM~\cite{bi2025prism}, as additional baselines. Since these additional baselines do not change the main conclusion and exhibit relatively weak performance, we defer their detailed results to Appendix~\ref{Benchmark_results}.


\paragraph{Evaluation}
\label{sec:evaluation}

We evaluate fine-tuned models on 12 benchmarks spanning general visual question answering, text-intensive visual understanding, document and chart understanding, and multimodal reasoning, including GQA~\cite{hudson2019gqa}, VizWiz~\cite{gurari2018vizwiz}, TextVQA~\cite{singh2019towards}, ScienceQA-IMG~\cite{lu2022learn}, MME~\cite{liang2024survey}, MMBench~\cite{liu2024mmbench}, AI2D~\cite{kembhavi2016diagram}, ChartQA~\cite{masry2022chartqa}, DocVQA~\cite{mathew2021docvqa}, InfoVQA~\cite{mathew2022infographicvqa}, MMStar~\cite{chen2024we}, and OCRBench~\cite{liu2024ocrbench}. Following COINCIDE and XMAS, we report Average Relative Performance (ARP) to normalize evaluation scales across datasets:
\begin{equation}
\text{ARP} = \frac{\text{Subset Data Performance}}{\text{Full Data Performance}} \times 100
.
\end{equation}

Detailed descriptions of all benchmarks and complete experimental results are provided in Appendix~\ref{Benchmark} and Appendix~\ref{Benchmark_results}, respectively.



\subsection{Main Results}
\label{sec:main_results}
\paragraph{Vision-Flan}

\begin{figure}[t]
    \centering

    \begin{subfigure}[t]{0.48\linewidth}
        \centering
        \includegraphics[
            width=\linewidth,
            height=0.25\textheight,
            keepaspectratio
        ]{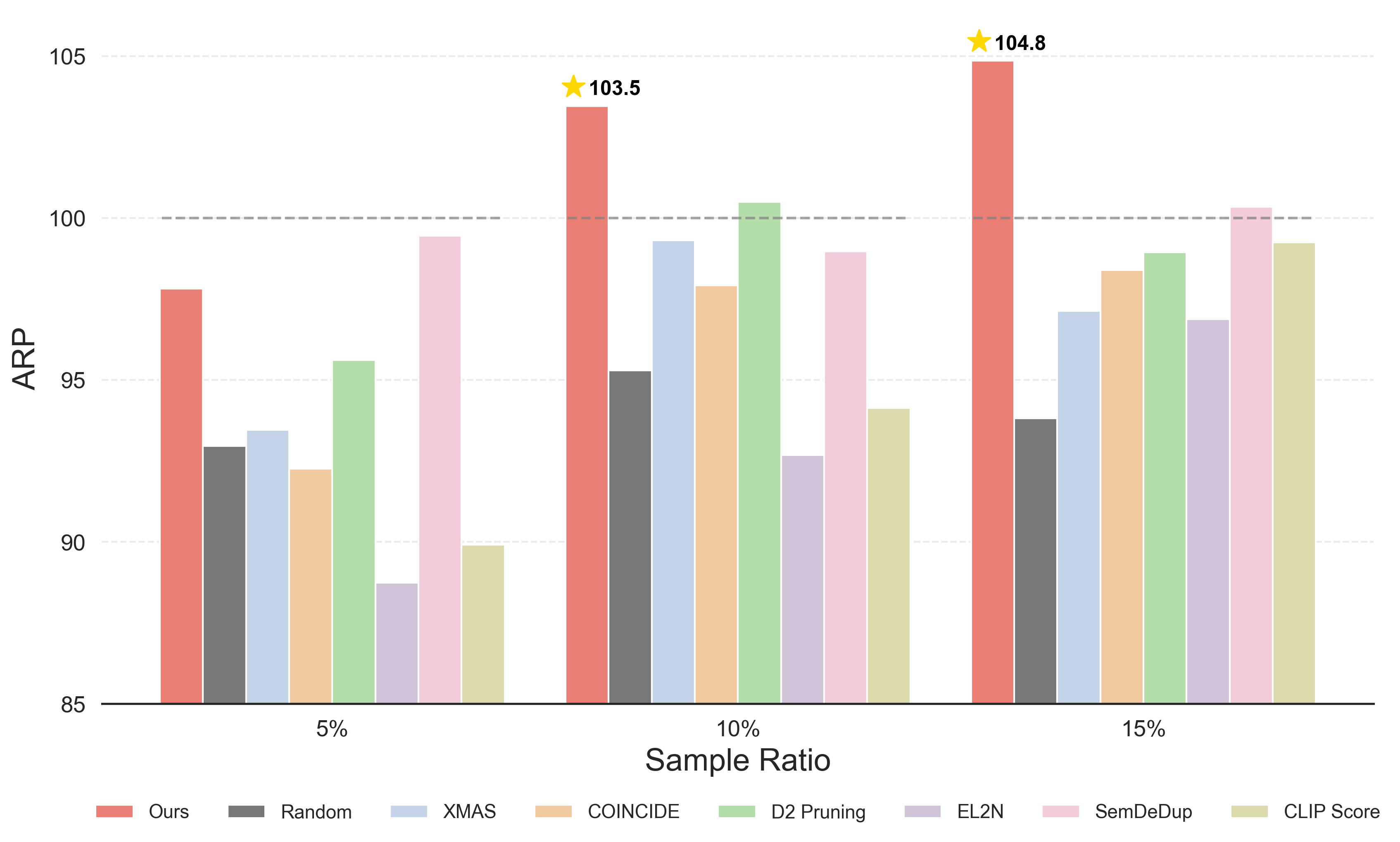}
        \caption{Vision-Flan.}
        \label{fig:data_scale_186k}
    \end{subfigure}
    \hfill
    \begin{subfigure}[t]{0.48\linewidth}
        \centering
        \includegraphics[
            width=\linewidth,
            height=0.25\textheight,
            keepaspectratio
        ]{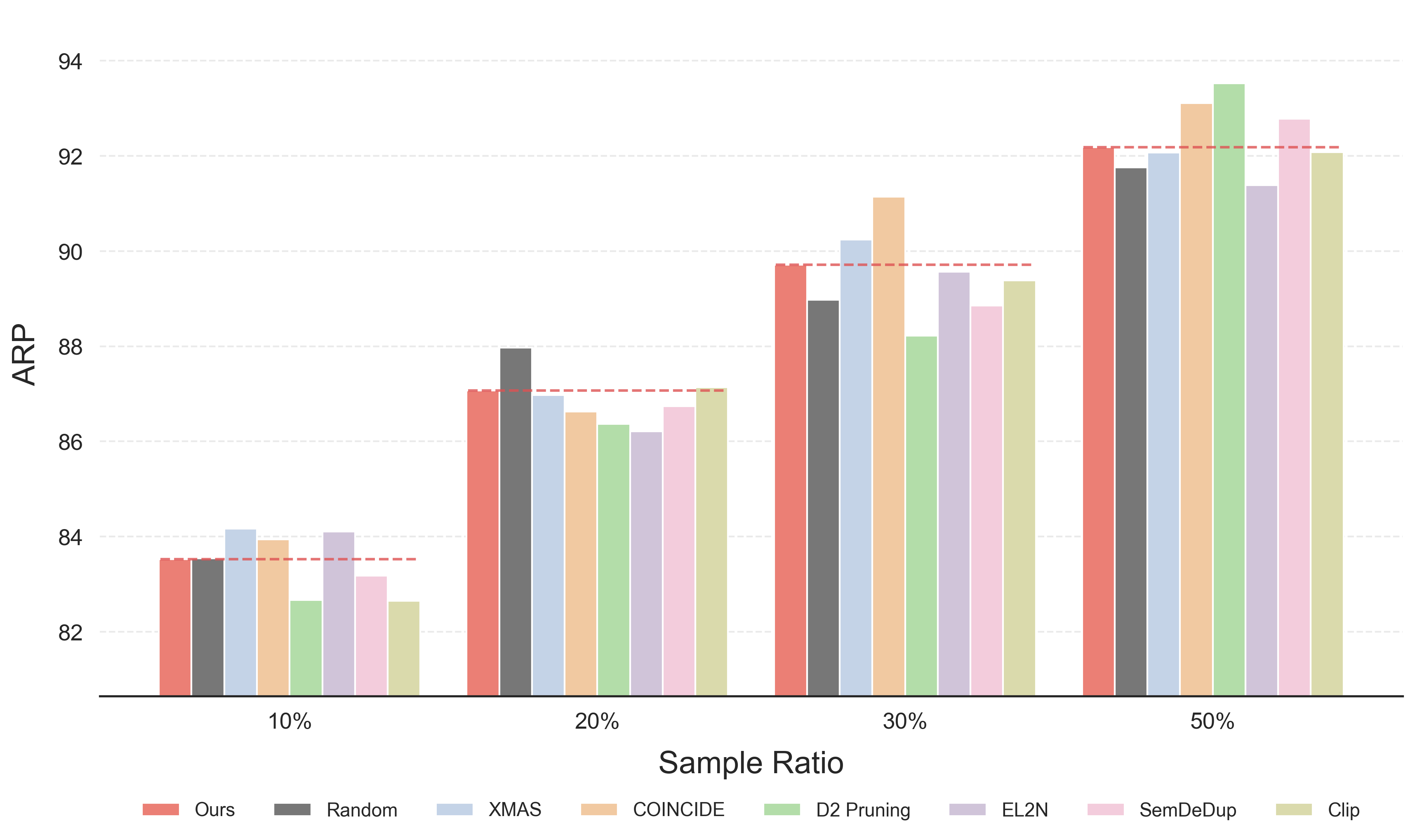}
        \caption{The Cauldron.}
        \label{fig:data_scale_100k}
    \end{subfigure}

    \caption{Performance comparison between CVS and baseline methods on Vision-Flan and The Cauldron. CVS performs best at 10\% and 15\% sampling ratios on Vision-Flan, even surpassing full-data training, and remains robust across sampling ratios on The Cauldron.}
    
    \label{fig:data_scale_ab}
\end{figure}

Due to its 187 diverse tasks, Vision-Flan poses substantial challenges for multi-task data selection. As shown in Figure~\ref{fig:data_scale_186k}, at a 5\% sampling ratio, CVS achieves the second-best performance, likely because its focus on near-boundary samples is less effective before a robust feature space is established under extremely limited data. Once the sampling ratio increases to 10\% and 15\%, this cold-start issue is mitigated, and the advantage of CVS becomes clear: it achieves state-of-the-art performance, significantly outperforming all baselines. In particular, CVS attains ARP of 103.5 and 104.8 at 10\% and 15\% sampling, respectively, surpassing even full-data fine-tuning (ARP = 100). These results strongly support our claim in Section~\ref{sec:filtering_protocol} that filtering out language-prior-dominated samples while prioritizing high-value, decision-boundary samples requiring vision-language joint reasoning leads to more effective gradient signals. Compared to XMAS and COINCIDE, CVS shows more robust performance advantages in the highly multi-task Vision-Flan setting, with the advantage increasing at medium and larger data budgets, indicating improved mitigation of task bias while maintaining broad visual coverage. Another noteworthy phenomenon is that baseline methods (e.g., XMAS and D2 Pruning) show large performance fluctuations as the sampling ratio increases, failing to achieve monotonic gains with larger data budgets. This instability suggests limited ability to distinguish informative samples from noise when expanding the subset. In contrast, CVS exhibits stable and consistent performance improvements across sampling ratios, indicating its robustness in selecting high-quality samples at different data scales. EL2N and CLIP-Score perform poorly across all sampling ratios, consistent with observations~\cite{lee2024coincide} that single-metric selection leads to biased and redundant subsets.

\paragraph{The Cauldron}
Compared with Vision-Flan, The Cauldron exhibits a substantially different form of supervision noise. Specifically, many of its visual instructions are generated via templated or rule-based procedures, leading to pronounced structural redundancy in the data distribution rather than noise dominated by linguistic shortcuts or semantic conflicts. This characteristic shifts the focus of data selection in a qualitatively different way and imposes distinct requirements on methods’ inductive biases. As shown in Figure~\ref{fig:data_scale_100k}, CVS shows stable and monotonic improvements across sampling ratios, indicating consistent and reliable selection across all budgets on The Cauldron with different noise types. However, at higher sampling ratios (e.g., 30\% and 50\%), some clustering-based methods (e.g., D2 Pruning and COINCIDE) slightly outperform CVS. We attribute this to the dominance of structural redundancy in The Cauldron: under larger data budgets, coverage and redundancy compression become increasingly influential, favoring clustering-based methods. In contrast, CVS focuses on identifying samples that provide discriminative signals for vision-language joint reasoning by suppressing semantic shortcuts and filtering semantically inconsistent data. As a result, when structural redundancy is the primary noise source, CVS is complementary to clustering- or deduplication-based methods rather than aiming for universal superiority across all budgets.

\section{Analysis and Ablation Studies}
\subsection{Effect of CVS Score Ranges}

\begin{figure}[tbp]
    \centering
    \includegraphics[width=0.85\linewidth]{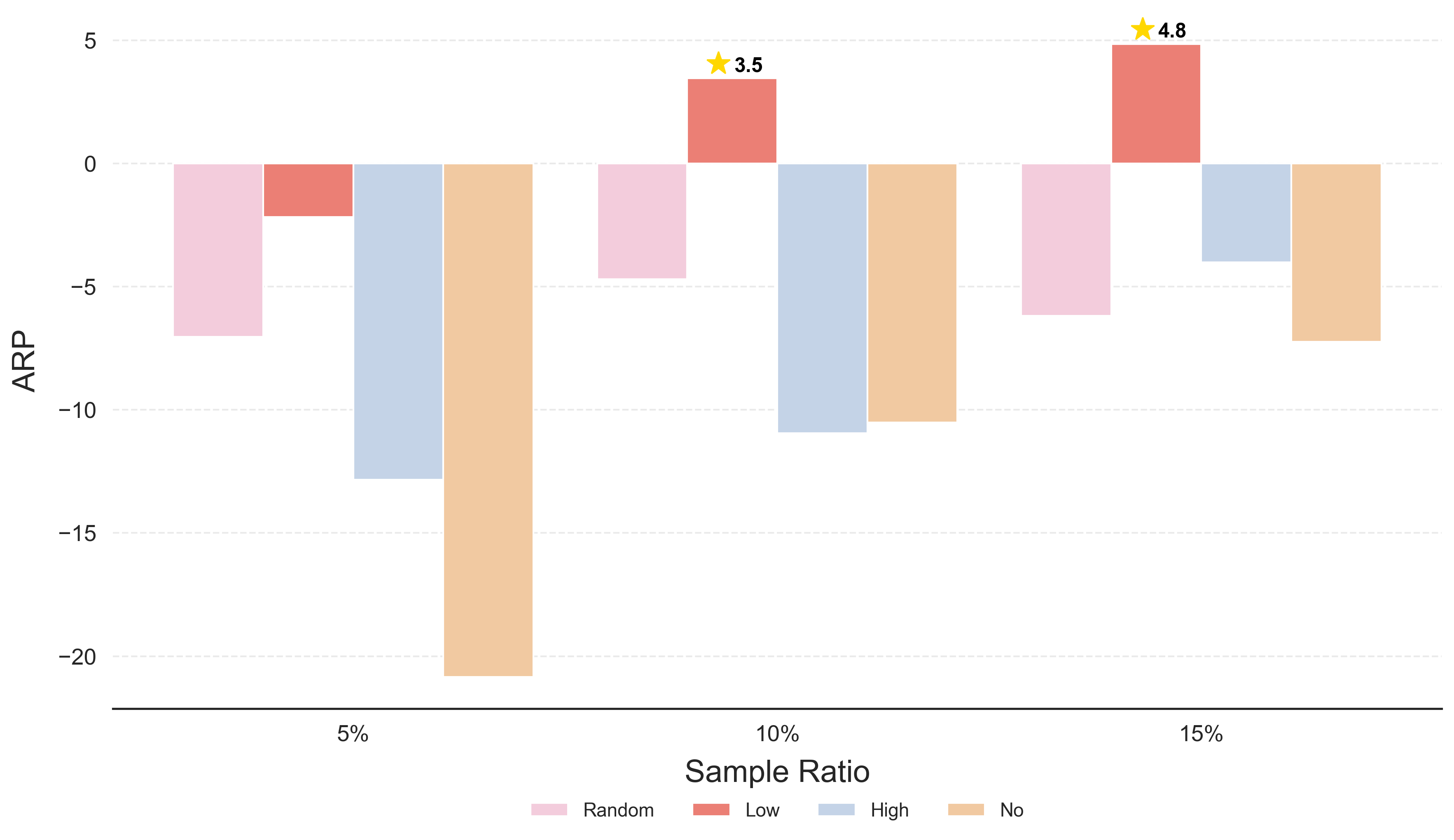}
    \caption{Effect of CVS score ranges on model performance. The low-score strategy consistently yields the largest gains across all budgets.}
    \label{fig:low_high_no}
\end{figure}

To study the impact of different CVS score ranges on model performance, we conduct a controlled comparison of sampling strategies on Vision-Flan using the pre-trained LLaVA-1.5-7B model. We compare Random sampling against three CVS-based variants: Low selects samples with the lowest $\text{CVS}_{\textsc{Yes}}$ from the filtered candidate pool, corresponding to samples near the decision boundary; High selects samples with the highest $\text{CVS}_{\textsc{Yes}}$ from the filtered candidate pool, representing easy high-confidence samples; and No selects samples with the highest $\text{CVS}_{\textsc{No}}$ over the entire dataset, corresponding to queries that strongly increase the rejection probability. The results in Figure~\ref{fig:low_high_no} show that sampling strategies lead to markedly different performance.



The Low strategy consistently outperforms the others, even under the cold-start setting with only 5\% data. As the sampling ratio increases to 10\% and 15\%, it becomes the only strategy that yields improvements over full-data fine-tuning, achieving gains of +3.5\% and +4.8\%, respectively. These results support our claim in Section~\ref{subsec:pre_for_hps} that samples near the decision boundary encourage vision-language joint reasoning and provide more informative optimization signals.


In contrast, the High strategy leads to clear performance degradation, performing significantly worse than random sampling at 10\% and 15\%. This also aligns with our claim in Section~\ref{subsec:pre_for_hps}: such samples exhibit strong linguistic correlations or common-sense shortcuts between the query $Q$ and answer $A$, allowing the model to predict while largely ignoring the image $I$. Training on these samples provides limited informational gain and wastes computational budget, ultimately harming generalization.

The No strategy yields the worst performance, consistently underperforming random sampling across all ratios, with especially severe degradation at 5\% and 10\%. High rejection shift typically reflects semantic misalignment or intrinsic noise (e.g., hallucinations), and forcing the model to fit such inconsistent samples disrupts the learned semantic space, leading to negative transfer.

Overall, the Low strategy maximizes data efficiency by identifying correct samples near the decision boundary. Representative Low, High, and No examples are provided in Appendix~\ref{cvsexample}.

\begin{figure}[tbp]
    \centering

    \begin{subfigure}[t]{0.45\linewidth}
        \centering
        \includegraphics[width=\linewidth]{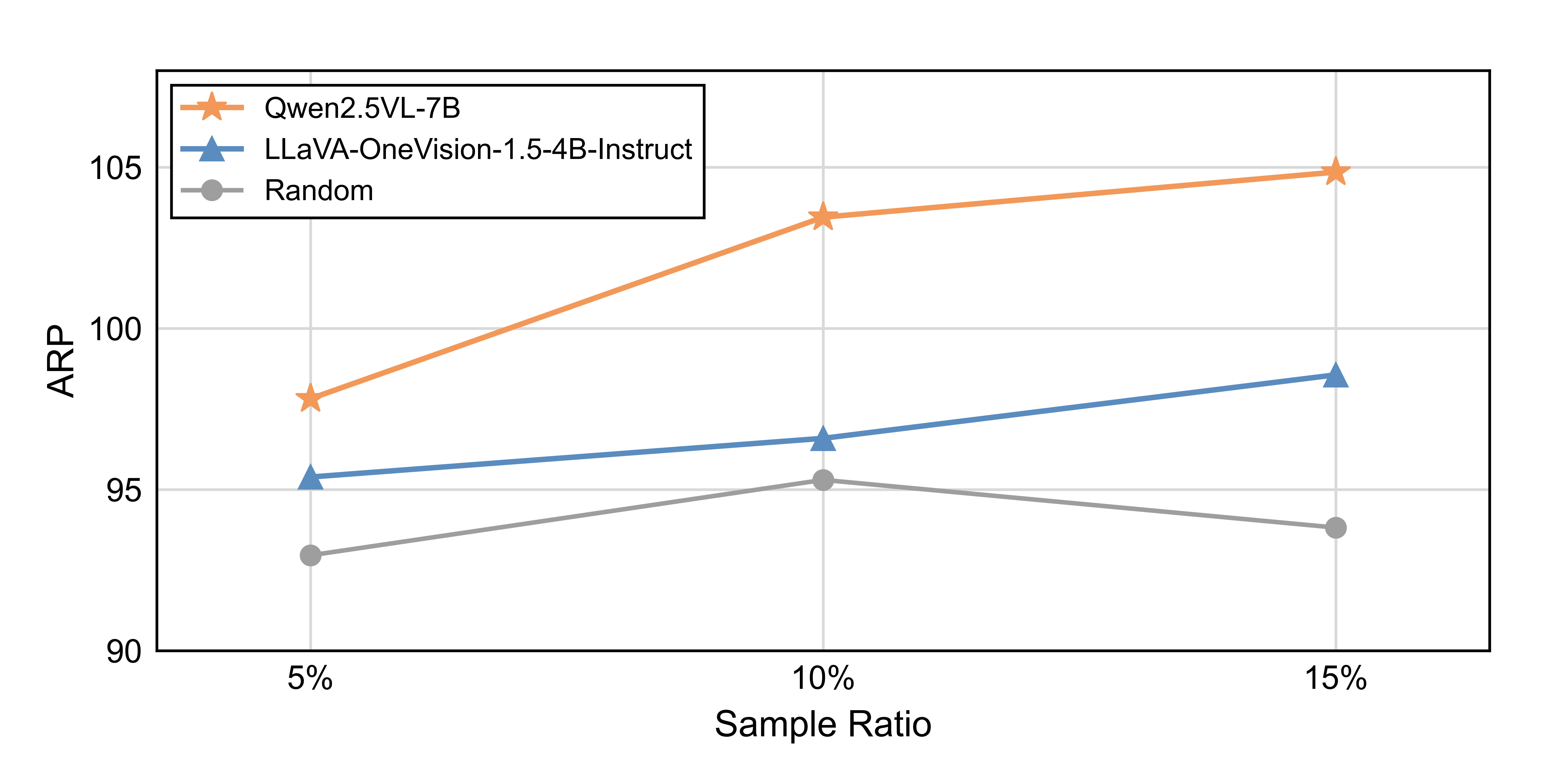}
        \caption{Evaluator architecture.}
        \label{fig:family}
    \end{subfigure}
    \hfill
    \begin{subfigure}[t]{0.45\linewidth}
        \centering
        \includegraphics[width=\linewidth]{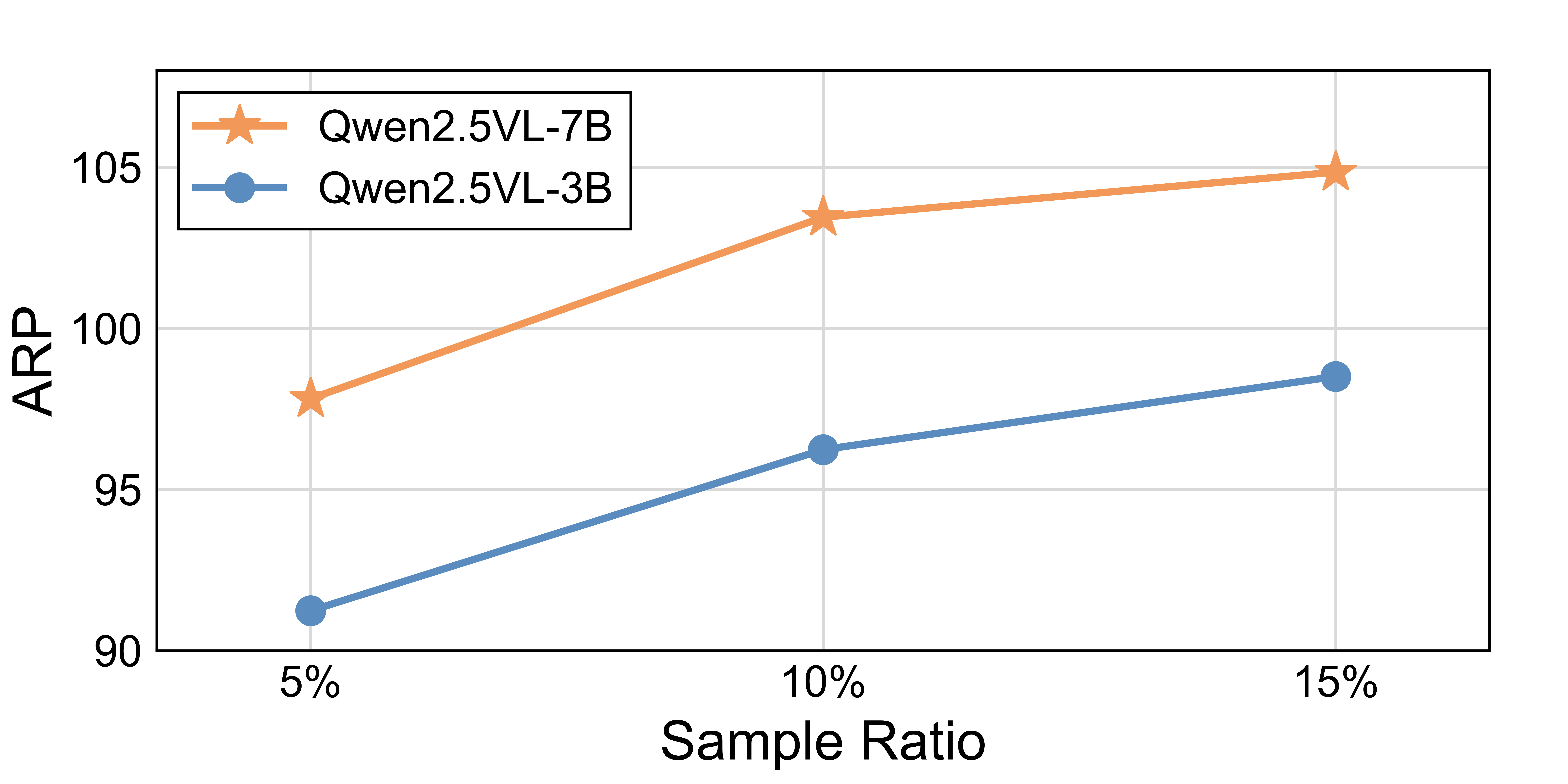}
        \caption{Evaluator scale.}
        \label{fig:param}
    \end{subfigure}

    \caption{Effect of evaluator architecture and scale on CVS performance.}
    \label{fig:evaluator_robustness}
\end{figure}

\subsection{Robustness to Evaluator}

This section evaluates the robustness of CVS with respect to evaluator architecture and evaluator scale. All experiments fix the target model to the pre-trained LLaVA-1.5-7B model and are conducted on Vision-Flan.

\textbf{Evaluator architecture} 
We replace the evaluator from Qwen2.5-VL-7B-Instruct with LLaVA-OneVision-1.5-4B-Instruct~\cite{an2025llava}, which belongs to a different model family. 
As shown in Figure~\ref{fig:family}, CVS consistently outperforms random sampling across all sampling ratios (5\%, 10\%, and 15\%) under both evaluator architectures, indicating that CVS does not rely on any specific model architecture. We further evaluate CVS with InternVL3-8B~\cite{zhu2025internvl3} as the evaluator and report the corresponding results in Appendix~\ref{Benchmark_results}.

\textbf{Evaluator scale} 
We further vary the evaluator size within the Qwen2.5-VL family, comparing 3B and 7B models. 
Figure~\ref{fig:param} shows a clear scaling trend: larger evaluators consistently lead to better downstream performance across all sampling ratios.

Overall, these results suggest that CVS benefits from stronger evaluators while remaining robust to evaluator architecture. This behavior contrasts with COINCIDE, which reports performance saturation or degradation when increasing proxy model size, indicating that CVS exhibits more favorable scalability. As the community continues to release increasingly powerful VLLMs, CVS can directly leverage stronger evaluators without modifying the algorithm.

\subsection{Robustness to Target Model}
\label{subsec:target_model}
To evaluate whether the selected data exhibits cross-model utility, we replace the target model from LLaVA-1.5-7B with the more capable pre-trained Qwen2-VL-2B~\cite{wang2024qwen2vlenhancingvisionlanguagemodels}, while fixing the evaluator as Qwen2-VL-7B-Instruct. All experiments are conducted on Vision-Flan. Given the higher capacity of Qwen2-VL-2B, we adopt a moderate learning rate of $1\mathrm{e}{-4}$, while keeping other hyperparameters unchanged. As shown in Table~\ref{tab:target}, CVS consistently outperforms random sampling across all sampling ratios (5\%, 10\%, and 15\%), even when applied to the stronger Qwen2-VL-2B target model. These results indicate that CVS captures model-agnostic data utility, selecting informative samples that remain effective across different target models. Similar results are observed when using LLaVA-1.5-13B as the target model (see Appendix~\ref{Benchmark_results}).

\subsection{Visual Anchoring}
CVS selects data by comparing $P(Y \mid I, Q, A)$ and $P(Y \mid I, A)$, where $Y \in \{\text{Yes}, \text{No}\}$. To assess the necessity of including image information in the denominator, we conduct an ablation study that removes $I$ from $P(Y \mid I, A)$, yielding a baseline conditioned only on the textual answer $A$. The experimental setup follows the main experiments and is conducted on Vision-Flan.

As shown in Table~\ref{tab:anchor}, data selected by CVS without visual anchoring consistently underperforms random sampling across all sampling ratios, with a performance gap exceeding 10 percentage points at the 5\% sampling rate. This result highlights the critical role of $P(Y \mid I, A)$ as a visual anchoring term in CVS. We analyze its effect from two perspectives.
\paragraph{Visual Anchoring Establishes a Reasoning Baseline}
The standard CVS formulation
\begin{equation}
    \frac{P(Y \mid I, Q, A)}{P(Y \mid I, A)}
\end{equation}
quantifies the information gain contributed by the query $Q$ under a fixed visual context, thereby identifying cases where visual evidence alone is insufficient and vision-language joint reasoning is required.

\paragraph{Semantic Collapse under the Degenerate Formulation}
When removing $I$ from the denominator, the formulation degenerates to
\begin{equation}
    \frac{P(Y \mid I, Q, A)}{P(Y \mid A)}.
\end{equation}

Under this degenerate setting, the score reflects to the mutual information between the multimodal input $(I, Q)$ and the textual prior of $A$. Applying Low strategy then favors samples where $(I, Q)$ contributes little to predicting $A$, resulting in weakly correlated image–text pairs or low-quality cases where the answer is independent of context (e.g., generic or uninformative text). In contrast, in the original CVS formulation, low positive scores have a different interpretation: they indicate that the query improves confidence in the correct answer in a non-trivial manner, requiring feature alignment grounded in visual priors rather than linguistic shortcuts. Consequently, retaining $I$ in the denominator is essential for CVS to distinguish low-value from high-value samples. Only by measuring the guiding effect of language on top of a visual perception baseline can CVS truly select data that promotes vision-language joint reasoning capability.
\subsection{Computation Time}
\label{sec:Computationtime}
The computational overhead of data selection methods critically affects their scalability. Table~\ref{tab:cost_comparison_symbols} compares the total GPU hours required by CVS with two recent state-of-the-art baselines, COINCIDE and XMAS, on The Cauldron. CVS demonstrates superior computational efficiency, processing the entire dataset in only 10.5 GPU hours, which saves 17.3\% and 44.4\% of the computation time compared to COINCIDE (12.7h) and XMAS (18.9h), respectively. This advantage mainly stems from the inference-only design of CVS. Unlike COINCIDE and XMAS, which require training a proxy model on the target dataset, CVS uses only forward inference with a frozen model. Eliminating proxy model training substantially reduces computational cost, enabling CVS to scale to multimodal datasets with millions or even hundreds of millions of samples.
\begin{table}[t!]
    \centering
    \scriptsize
    \captionsetup{skip=5pt}

    \begin{minipage}[t]{0.31\textwidth}
        \vspace{0pt}
        \centering
        \captionof{table}{Performance under stronger target model.}
        \label{tab:target}
        \vspace{0.75em}
        \setlength{\tabcolsep}{4pt}
        \begin{tabular}{lccc}
            \toprule
            Method & 5\% & 10\% & 15\% \\
            \midrule
            Random & 98.34 & 98.64 & 97.80 \\
            CVS & \textbf{99.65} & \textbf{100.09} & \textbf{99.65} \\
            \bottomrule
        \end{tabular}
    \end{minipage}
    \hfill
    \begin{minipage}[t]{0.31\textwidth}
        \vspace{0pt}
        \centering
        \captionof{table}{Effect of removing visual anchoring.}
        \label{tab:anchor}
        \vspace{0.75em}
        \setlength{\tabcolsep}{4pt}
        \begin{tabular}{lccc}
            \toprule
            Method & 5\% & 10\% & 15\% \\
            \midrule
            Random & \textbf{92.96} & \textbf{95.30} & \textbf{93.62} \\
            CVS & 82.44 & 89.48 & 88.44 \\
            \bottomrule
        \end{tabular}
    \end{minipage}
    \hfill
    \begin{minipage}[t]{0.31\textwidth}
        \vspace{0pt}
        \centering
        \captionof{table}{Computation time comparison on The Cauldron.}
        \label{tab:cost_comparison_symbols}
        \setlength{\tabcolsep}{4pt}
        \begin{tabular}{lcc}
            \toprule
            Method & Training-Free & GPU Hours \\
            \midrule
            XMAS & \xmark & 18.9 \\
            COINCIDE & \xmark & 12.7 \\
            CVS & \cmark & \textbf{10.5} \\
            \bottomrule
        \end{tabular}
    \end{minipage}
\end{table}

\section{Conclusion and Limitations}
\label{sec:conclusion_limitations}

\paragraph{Conclusion}
In this paper, we propose CVS, a training-free data selection method. CVS is based on the observation that high-value multimodal samples should require the question to meaningfully change the model's judgment of answer validity given the image. By comparing a frozen VLLM's verdicts with and without conditioning on the question, CVS filters semantically inconsistent samples and prioritizes informative hard positives that better encourage vision-language joint reasoning. Experiments on Vision-Flan and The Cauldron demonstrate that CVS achieves strong data efficiency, competitive downstream performance, and lower computational cost compared with recent baselines.

\paragraph{Limitations}
Despite its effectiveness, CVS has several limitations. First, although CVS is robust across different evaluators, its performance may degrade when the frozen evaluator has strong biases or poor calibration. Second, while CVS avoids proxy-model training, it still requires inference over the candidate pool, leaving room for further efficiency improvements. Third, CVS mainly addresses semantic conflicts and language-prior shortcuts, and may be less effective when structural redundancy is the primary source of data inefficiency. Future work can further combine CVS with diversity-based methods to improve data selection under heterogeneous data distributions.

\newpage
\bibliographystyle{plainnat}
\bibliography{references}

\newpage

\newpage
\appendix


\section{Related Work}
\label{sec:related_work}

Existing VLLM data selection methods can be categorized into score-based and clustering-based methods. In this section, we present a detailed overview of these methods.
\subsection{Score-based Methods}

Score-based methods assign an importance score to each sample and perform selection accordingly. Early methods rely on simple proxies such as prediction error or perplexity (e.g., EL2N~\cite{paul2021deep}, MP~\cite{marion2023moreinvestigatingdatapruning}). More recent methods incorporate gradient-based signals to estimate sample contribution or influence, including Self-Filter~\cite{chen2024visionlanguagemodelstrongfilter}, LESS~\cite{xia2024lessselectinginfluentialdata}, ICONS~\cite{wu2025iconsinfluenceconsensusvisionlanguage}, and TIVE~\cite{liu2025less}. Other variants further consider diversity or grouped sampling (e.g., COIDO~\cite{yan2025coidoefficientdataselection}, MLLM-Selector~\cite{ma2025mllmselectornecessitydiversitydrivenhighvalue}). Most score-based methods require training proxy models, incurring substantial overhead for large-scale multimodal data. Training-free alternatives such as CLIP-Score~\cite{hessel2021clipscore} avoid this cost but are limited to coarse-grained semantic alignment~\cite{nam2025extract}. In contrast, our method leverages a frozen VLLM to model conditional shifts via relative judgment changes, enabling fine-grained semantic relationships modeling without proxy training.

\subsection{Clustering-based Methods}
Clustering-based methods promote diversity by modeling relationships among samples, typically by performing clustering in the embedding space of a specific model layer. Early works such as Self-Sup~\cite{sorscher2022beyond} and SemDeDup~\cite{abbas2023semdedupdataefficientlearningwebscale} remove redundant samples based on cluster structure or intra-cluster similarity, while D2 Pruning~\cite{maharana2023d2pruningmessagepassing} models soft neighborhoods via graph message passing. More recent methods exploit richer or cross-modal representations. COINCIDE~\cite{lee2024coincide} clusters joint representations from multiple layers to capture concept--skill structure, and XMAS~\cite{naharas2025dataselectionfinetuningvision} clusters samples based on cross-modal attention trajectories. ARDS~\cite{yang2025data} further improves robustness through worst-case subgroup construction and weighted sampling. Existing clustering-based methods typically involve additional operations after feature extraction, and some still suffer from the overhead of training proxy models. In contrast, our method is more cost-efficient, as it directly leverages a frozen VLLM without requiring any additional post-clustering procedures or proxy training.

\section{Distributional Perspectives on Data Selection}
\label{sec:preliminaries}

We emphasize that this perspective is introduced as an interpretative lens, rather than a formal objective or a theoretical derivation of our method. In practice, training data are rarely sampled directly from the real-world distribution $\mathcal{P}_{\text{real}}$, but instead from a constructed distribution $\mathcal{P}_{\text{train}}$ formed through data integration or filtering. Building on recent advances in data selection~\cite{dohmatob2025sometimestheorydatacuration}, the training of VLLMs can be viewed as an interaction among three components:
\begin{itemize}
    \item \textbf{Generator ($\mathcal{G}_\theta$):} The VLLM learner, whose parameters $\theta$ are optimized via gradient descent to model conditional distributions.
    
    \item \textbf{Evaluator ($\mathcal{D}_\phi$):} A parameterized model or scoring function that selects or reweights training samples from raw data sources.
    
    \item \textbf{Target Distribution ($\mathcal{P}_{real}$):} The real-world distribution used to assess generalization performance.
\end{itemize}

Under this formulation, the generalization error $L_{\text{test}}$ depends not only on the training set size $N$, but also on the interaction between the Generator’s modeling capacity and the Evaluator’s ability to align $\mathcal{P}{\text{train}}$ with $\mathcal{P}{\text{real}}$. Prior theory suggests a phase transition: when the Generator is sufficiently expressive and the Evaluator effectively filters for informative, distribution-consistent samples, strong generalization can be achieved with substantially reduced data, corresponding to a ``less is more'' regime.

\section{CVS Prompt Templates}
\label{Templates}

\begin{figure}
    \centering
    
    \begin{subfigure}{0.9\linewidth} 
        \centering
        \includegraphics[width=0.85\linewidth]{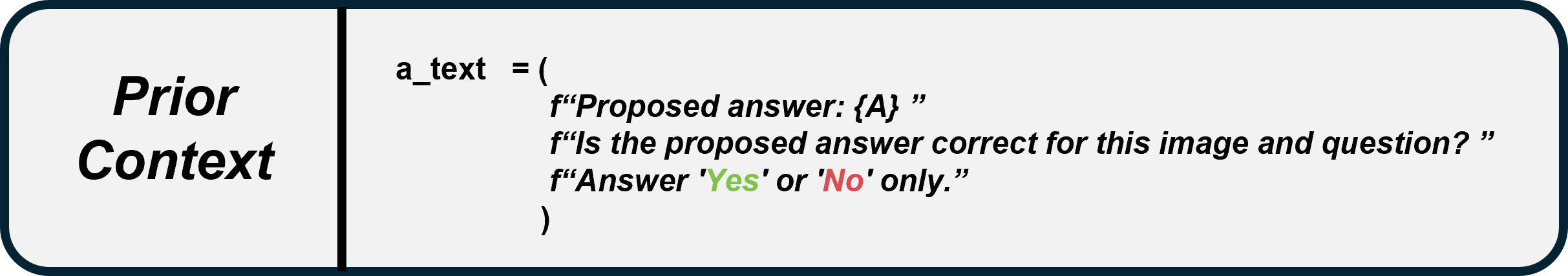}
        \caption{Prior Context Prompt.} 
        \label{fig:a}
    \end{subfigure}
    
    \vspace{1em} 
    
    \begin{subfigure}{0.9\linewidth}
        \centering
        \includegraphics[width=0.85\linewidth]{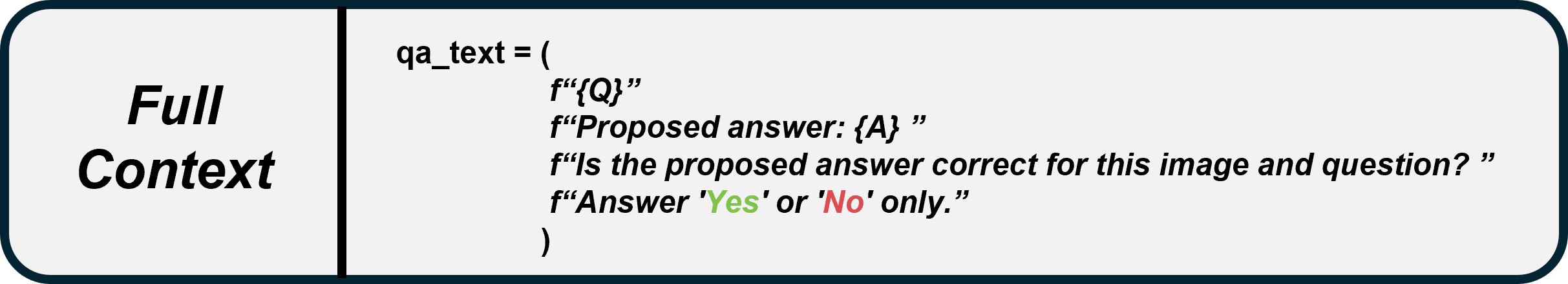}
        \caption{Full Context Prompt.} 
        \label{fig:qa}
    \end{subfigure}
    
    \caption{Prompt templates for CVS computation.}
    \label{fig:cvs_prompt}
\end{figure}

As illustrated in Figure~\ref{fig:cvs_prompt}, we provide the complete prompt templates used to obtain binary answer validity judgments from a frozen VLLM. The model is instructed to output only ``Yes'' or ``No''.

\section{Complementarity of Affirmation and Rejection Shifts}
\label{app:yes_no_complementarity}

In the main paper, we use Conditional Affirmation Shift ($\text{CVS}_{\textsc{Yes}}$) and Conditional Rejection Shift ($\text{CVS}_{\textsc{No}}$) as two separate constraints. Since the evaluator predicts over $\{\text{Yes}, \text{No}\}$, one may wonder whether these two quantities are redundant, and whether they can be replaced by a single coupled score. We therefore conduct an ablation on Vision-Flan under the 10\% budget setting to examine the necessity of keeping them separate.

Specifically, we compare three types of variants: using only $\text{CVS}_{\textsc{Yes}}$, using only $\text{CVS}_{\textsc{No}}$, and replacing the two constraints with a coupled log-odds change:
\begin{equation}
\Delta =
\log \frac{P(\text{Yes}\mid I,Q,A)}{P(\text{No}\mid I,Q,A)}
-
\log \frac{P(\text{Yes}\mid I,A)}{P(\text{No}\mid I,A)}.
\end{equation}
For the coupled score, we evaluate three selection strategies, denoted as $\Delta_{P_{\text{Min}}}$, $\Delta_{P_{\text{Random}}}$, and $\Delta_{P_{\text{Max}}}$.

As shown in Table~\ref{tab:yes_no_complementarity}, both individual constraints provide useful signals, with $\text{CVS}_{\textsc{Yes}}$ and $\text{CVS}_{\textsc{No}}$ achieving 100.59 and 101.89 ARP, respectively. However, combining the two shifts achieves the best performance, improving the ARP to 103.45. In contrast, the coupled log-odds variants are consistently inferior to the separated formulation, with the best coupled variant reaching only 102.11 ARP. These results suggest that although the Yes/No probabilities are naturally correlated, their shifts are not redundant in practice. Separating $\text{CVS}_{\textsc{Yes}}$ and $\text{CVS}_{\textsc{No}}$ better preserves the distinction between answer support and semantic conflict, which leads to more effective filtering than collapsing them into a single scalar.

\section{Prompt Sensitivity Analysis}
\label{app:prompt_sensitivity}

CVS relies on evaluator probabilities assigned to binary answer tokens. Therefore, the resulting scores may be affected by the choice of answer verbalizers or by the specific wording of the evaluator prompt. To examine the robustness of CVS to these design choices, we conduct a prompt sensitivity analysis on Vision-Flan under the 10\% data budget.

We consider two types of variants. First, we keep the prompt template unchanged and vary only the binary answer tokens, replacing $\{\textsc{Yes}, \textsc{No}\}$ with $\{\textsc{Correct}, \textsc{Incorrect}\}$ and $\{\textsc{Right}, \textsc{Wrong}\}$. Second, we keep the answer tokens as $\{\textsc{Yes}, \textsc{No}\}$ but modify the prompt wording by replacing the original template with the question ``Can the proposed answer be regarded as correct for this image and question?'' All other components of the selection pipeline are kept unchanged.

As shown in Table~\ref{tab:prompt_sensitivity}, all CVS variants substantially outperform random selection. More importantly, changing the answer verbalizers leads to only minor performance variation: $\{\textsc{Yes}, \textsc{No}\}$, $\{\textsc{Correct}, \textsc{Incorrect}\}$, and $\{\textsc{Right}, \textsc{Wrong}\}$ achieve 103.45, 103.69, and 103.00 ARP, respectively. The alternative prompt template further achieves 104.54 ARP. These results indicate that CVS is not overly sensitive to a particular choice of binary answer tokens or prompt wording. Instead, its effectiveness mainly comes from the conditional probability shift induced by the image--question--answer context, rather than from a specific prompt instantiation.

\begin{table*}[t]
\centering

\begin{minipage}[t]{0.48\textwidth}
    \centering
    \caption{Effect of separating affirmation and rejection constraints.}
    \label{tab:yes_no_complementarity}
    \setlength{\tabcolsep}{10pt}
    \begin{tabular}{lc}
        \toprule
        Score & ARP \\
        \midrule
        $\text{CVS}_{\textsc{Yes}}$ & 100.59 \\
        $\text{CVS}_{\textsc{No}}$ & 101.89 \\
        $\Delta_{P_{\text{Min}}}$ & 94.53 \\
        $\Delta_{P_{\text{Random}}}$ & 101.62 \\
        $\Delta_{P_{\text{Max}}}$ & 102.11 \\
        $\text{CVS}_{\textsc{Yes}} + \text{CVS}_{\textsc{No}}$ & \textbf{103.45} \\
        \bottomrule
    \end{tabular}
\end{minipage}
\hfill
\begin{minipage}[t]{0.48\textwidth}
    \centering
    \caption{Prompt sensitivity analysis of CVS.}
    \label{tab:prompt_sensitivity}
    \vspace{18pt}
    \setlength{\tabcolsep}{10pt}
    \begin{tabular}{lc}
        \toprule
        Method & ARP \\
        \midrule
        Random & 95.30 \\
        $\{\textsc{Yes}, \textsc{No}\}$ & 103.45 \\
        $\{\textsc{Correct}, \textsc{Incorrect}\}$ & 103.69 \\
        $\{\textsc{Right}, \textsc{Wrong}\}$ & 103.00 \\
        Can...question? & \textbf{104.54} \\
        \bottomrule
    \end{tabular}
\end{minipage}

\end{table*}

\section{Dataset Details}
\label{Dataset_dtl}
\subsection{Vision-Flan}

Vision-Flan~\cite{xu2024visionflanscalinghumanlabeledtasks} is a high-quality human-annotated visual instruction tuning dataset designed to systematically enhance the generalization capability of vision-language models across diverse visual tasks. The dataset is constructed entirely from publicly available academic vision datasets and covers 187 finely grained visual tasks, each accompanied by expert-written and rigorously validated task instructions. This design significantly reduces synthetic bias and the risk of hallucination.

Prior studies have demonstrated that Vision-Flan offers substantial benefits in improving overall visual understanding while mitigating catastrophic forgetting, making it one of the publicly available VIT datasets with the widest task diversity and highest annotation reliability to date. Each task contains approximately 1,000 samples, resulting in a total of 186K samples, all of which are used in our experiments.

\subsection{The Cauldron}

The Cauldron~\cite{laurenccon2024matters} is a large-scale, high-quality visual instruction tuning dataset specifically constructed for instruction tuning of general-purpose vision-language large models. It integrates 50 vision-language datasets spanning a wide range of task types, including visual question answering, image captioning, text transcription, document and table understanding, chart reasoning, visual logical reasoning, and code generation across diverse and complex scenarios.

All constituent datasets are unified into a standardized question–answer format, and rigorous deduplication and evaluation-set decontamination procedures are applied during dataset construction to effectively mitigate the risk of data leakage. Given the substantial scale of the Cauldron dataset, to facilitate broad and fair evaluation of multiple baseline methods under a reasonable computational budget, we construct a representative subset of 100K samples by random sampling, while strictly preserving the original task distribution.

\section{Hyperparameter Settings for Baseline Methods}
\label{Baseline_hyperparam}
For all baseline methods other than XMAS, we use the hyperparameter settings reported in the COINCIDE paper. For XMAS, we follow the hyperparameter configuration described in its paper, as summarized in Table~\ref{tab:hyper_settings}.
\begin{table}
    \centering
    \caption{Hyperparameter configurations.}
    \label{tab:hyper_settings}
    \begin{tabular}{ll}
        \toprule
        Method & Settings \\ 
        \midrule
        CLIP-Score      & high score selected \\
        EL2N            & medium score selected \\
        SemDeDup        & $K: 5,000$ \\
        D2 Pruning      & $k: 5, \gamma_r: 0.4, \gamma_f: 1.0$ \\
        COINCIDE        & $K: 5,000, \tau: 0.1$ \\
        XMAS            & $K: 1,000, T: 7$ \\
        \bottomrule
    \end{tabular}
\end{table}

\section{Benchmark Details}
\label{Benchmark}
\paragraph{GQA}
GQA~\cite{hudson2019gqa} is a visual question answering benchmark designed for real-world visual reasoning. It automatically generates questions with explicit compositional structure from scene graphs and provides explicit annotations of multi-step reasoning processes. By systematically balancing the distributions of questions and answers, GQA substantially reduces language bias, enabling more reliable evaluation of models’ compositional vision-language reasoning capabilities.

\paragraph{VizWiz}
VizWiz~\cite{gurari2018vizwiz} is a visual question answering benchmark targeting real-world assistive scenarios, with data collected from images captured by blind users using mobile phones along with their spoken questions. The benchmark contains a large number of low-quality images and unanswerable questions, and explicitly incorporates answerability prediction into its evaluation protocol, enabling assessment of VQA models’ robustness in real-world, unconstrained environments.

\paragraph{TextVQA}
TextVQA~\cite{singh2019towards} is a visual question answering benchmark focused on understanding text in images, comprising approximately 28K natural scene images and 45K questions that require reading and reasoning over textual content to answer. The benchmark emphasizes open-vocabulary answers and OCR-based copying mechanisms, and is designed to evaluate models’ text reading and multimodal reasoning abilities in real-world scenarios.

\paragraph{ScienceQA-IMG}
ScienceQA-IMG~\cite{lu2022learn} is a subset of ScienceQA that contains only questions with image-based context, including both natural scene images and instructional diagrams. It requires models to perform multiple-choice scientific question answering by integrating visual information. This subset emphasizes image understanding and multi-step scientific reasoning, and provides ground-truth answers along with corresponding explanation annotations, making it suitable for evaluating multimodal reasoning performance.

\paragraph{MME}
MME~\cite{liang2024survey} is a comprehensive benchmark designed to systematically assess the perception and reasoning capabilities of multimodal large language models. It covers a diverse set of evaluation dimensions, including object recognition, attribute understanding, spatial reasoning, OCR, commonsense reasoning, and visual reasoning, spanning both perceptual-level and cognitive-level skills. MME adopts a unified evaluation protocol with carefully curated prompts and scoring criteria, enabling fine-grained and reliable comparison of multimodal models across heterogeneous abilities.

\paragraph{MMBench}
MMBench~\cite{liu2024mmbench} is a multi-capability evaluation benchmark for large-scale vision-language models, consisting of approximately 3K multiple-choice questions that cover 20 fine-grained ability dimensions, spanning both perceptual and reasoning skills. The benchmark employs strict data quality control and a CircularEval evaluation strategy, enabling more robust and reliable comparison of models’ comprehensive multimodal capabilities.

\paragraph{AI2D}
AI2D~\cite{kembhavi2016diagram} is a benchmark for scientific diagram understanding, consisting of over 5K diagrams drawn from elementary-level science textbooks and approximately 15K multiple-choice questions. The benchmark emphasizes modeling entities, relations, and structural layouts within diagrams, and is designed to evaluate models’ diagram parsing and relation-based reasoning abilities.

\paragraph{ChartQA}
ChartQA~\cite{masry2022chartqa} is a question answering benchmark for real-world chart understanding, featuring charts derived from real data sources and human-authored questions that emphasize joint visual and logical reasoning. The benchmark supports open-vocabulary answers and requires models to simultaneously comprehend visual chart elements and underlying numerical relationships, making it suitable for evaluating complex chart reasoning capabilities.

\paragraph{DocVQA}
DocVQA~\cite{mathew2021docvqa} is a visual question answering benchmark for document image understanding, comprising approximately 12K real-world document images and 50K text-based questions. The benchmark emphasizes joint modeling of document layout, structure, and dense textual content, and is designed to evaluate models’ document-level reading comprehension and information localization abilities.

\paragraph{InfoVQA}
InfoVQA~\cite{mathew2022infographicvqa} is a visual question answering benchmark for infographic understanding, comprising approximately 5.4K infographic images and 30K questions that require joint modeling of text, layout, graphics, and data visualization elements. The benchmark emphasizes reasoning skills such as counting, ordering, and basic arithmetic, and is designed to evaluate models’ performance on complex document-level vision-language understanding tasks.

\paragraph{MMStar}
MMStar~\cite{chen2024we} is a multimodal evaluation benchmark that emphasizes visual indispensability, consisting of 1,500 carefully human-filtered questions for which correct answers must rely on visual information. The benchmark spans 6 core capabilities and 18 fine-grained dimensions, and is designed to more faithfully evaluate large models’ multimodal understanding and reasoning abilities, while explicitly reducing the impact of textual shortcuts and data leakage.

\paragraph{OCRBench}
OCRBench~\cite{liu2024ocrbench} is a comprehensive benchmark for evaluating the OCR capabilities of large-scale multimodal models, covering five task categories: text recognition, scene text VQA, document VQA, key information extraction, and handwritten formula recognition. The benchmark consists of 1,000 human-curated and manually verified question–answer pairs, and is designed to systematically analyze models’ actual strengths and limitations across diverse text-centric visual tasks.

\section{Benchmark Result Details}
\label{Benchmark_results}
This section summarizes the detailed results of all experiments in this paper.
Table~\ref{tab:full_comparison_vf} reports the complete results on Vision-Flan for different data selection methods using 5\%, 10\%, and 15\% data subsets.
Table~\ref{tab:full_comparison_cr} presents the complete results on The Cauldron for different data selection methods using 10\%, 20\%, 30\%, and 50\% data subsets.
Table~\ref{tab:full_comparison_lhn} shows the impact of different CVS score selection strategies (Low / High / No) on model performance on Vision-Flan.
Tables~\ref{tab:full_comparison_arch} and~\ref{tab:as_full_comparison} compare the performance of CVS under different evaluator architectures and different evaluator scale settings, respectively.
Table~\ref{tab:tm_full_comparison} and~\ref{tab:tm_13_full_comparison} report robustness evaluation results when changing the target model while keeping the evaluator fixed.
Finally, Table~\ref{tab:va_full_comparison} presents an ablation study that removes the visual anchoring term, aiming to analyze the role of visual information in the design of CVS.

\section{Examples across CVS Score Ranges}
\label{cvsexample}
\begin{figure}[p]
    \centering
    \includegraphics[width=0.7\linewidth]{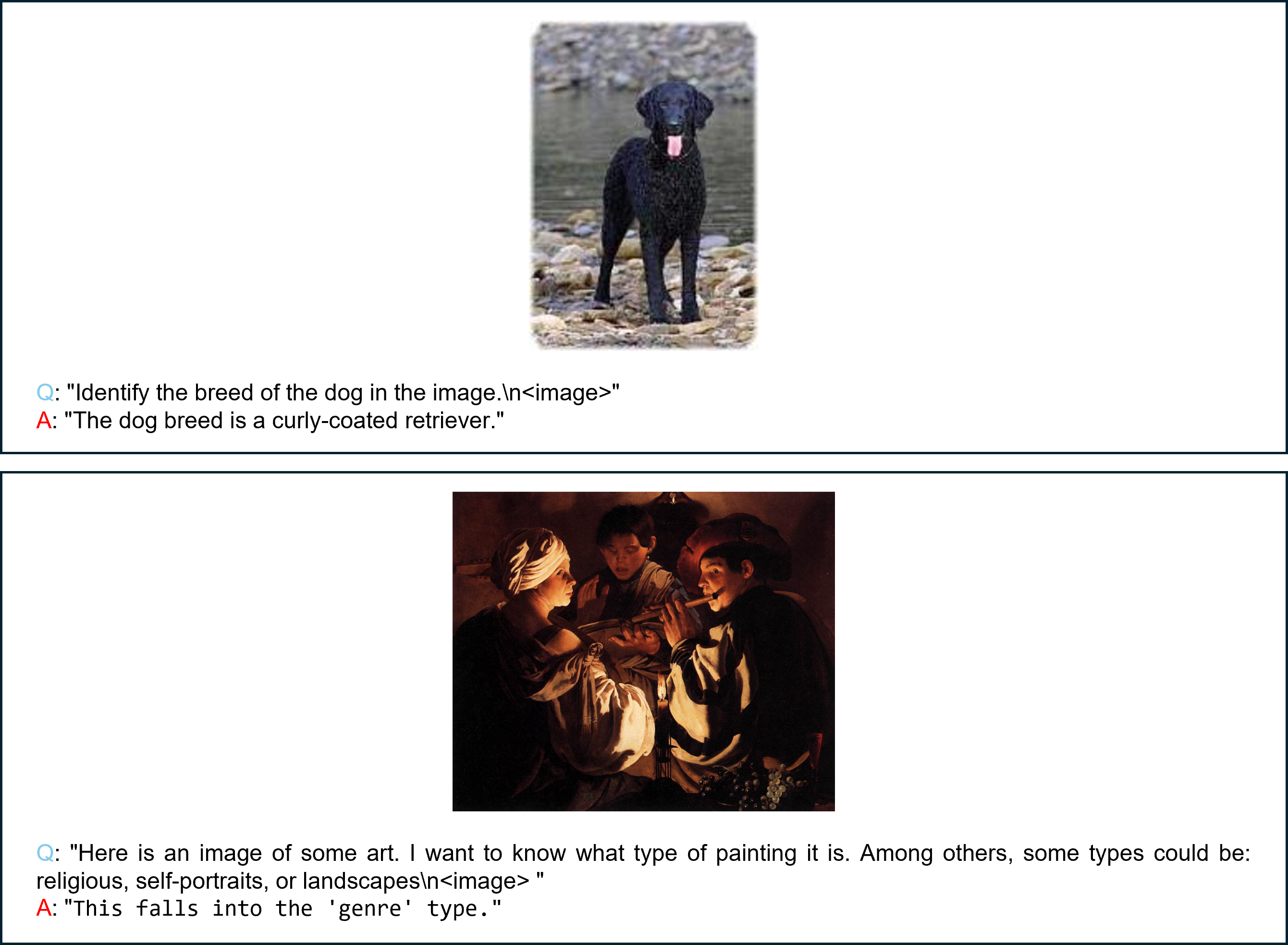}
    \caption{Representative examples selected by the Low strategy.}
    \label{fig:lowexample}
\end{figure}

\begin{figure}[p]
    \centering
    \includegraphics[width=0.7\linewidth]{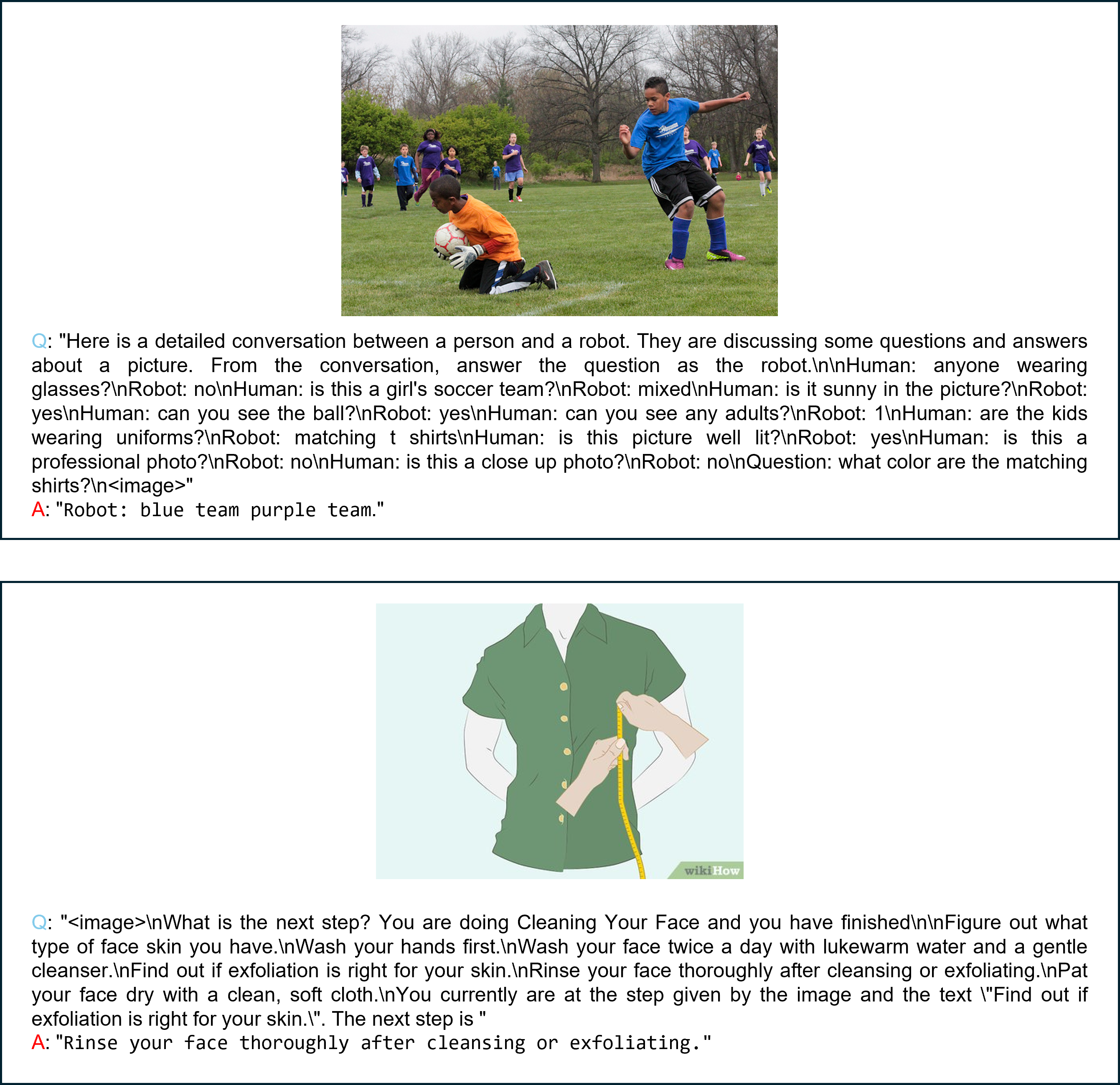}
    \caption{Representative examples selected by the High strategy.}
    \label{fig:highexample}
\end{figure}

\begin{figure}
    \centering
    \includegraphics[width=0.7\linewidth]{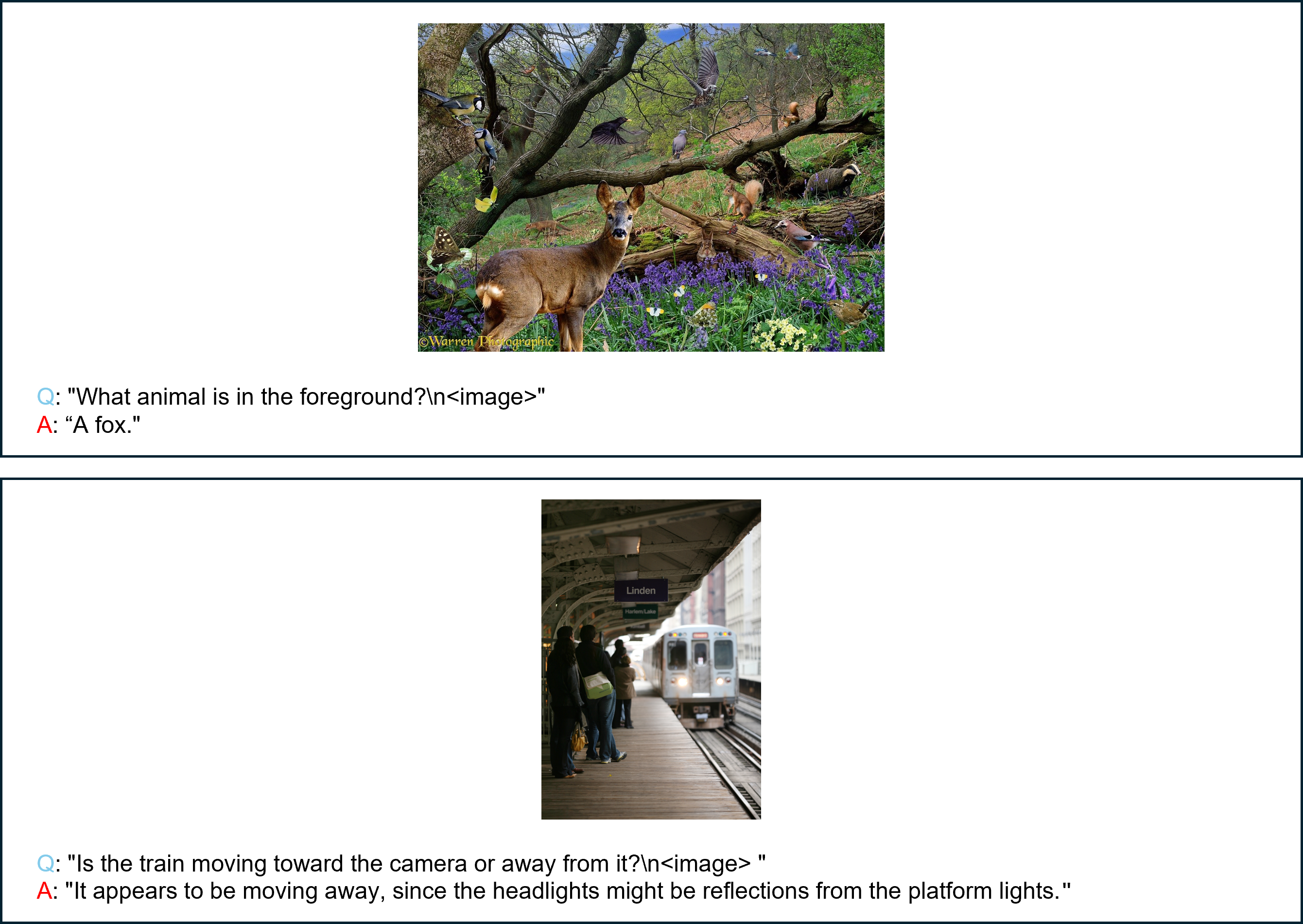}
    \caption{Representative examples selected by the No strategy.}
    \label{fig:noexample}
\end{figure}

In this section, we provide representative examples across different CVS score ranges, including Low, High, and No samples. The corresponding examples are shown in Figures~\ref{fig:lowexample}, \ref{fig:highexample}, and \ref{fig:noexample}, respectively, to qualitatively illustrate the characteristics of samples within each score range.

\section{Discussion on Absolute Verdict Confidence}

CVS primarily relies on relative verdict shifts rather than directly ranking samples by the absolute probability $P(\mathrm{Yes}\mid I,Q,A)$. This design does not imply that the absolute verdict confidence is unimportant. In principle, $P(\mathrm{Yes}\mid I,Q,A)$ reflects the evaluator's confidence in the correctness of the answer under the full multimodal context, and extremely low confidence may indicate ambiguous, noisy, or low-quality supervision. However, we find that absolute confidence values are less reliable as a universal selection criterion across heterogeneous visual instruction tasks, since different tasks, answer formats, and prompt styles can induce substantially different confidence ranges in the evaluator. As a result, a fixed absolute-confidence threshold may either over-filter valid but challenging samples or retain easy samples with high linguistic priors.

In contrast, CVS measures the conditional change in the evaluator's judgment caused by introducing the question $Q$ under the same visual context. This relative formulation provides a more task-stable signal of whether the question contributes useful information for validating the answer. Moreover, low-quality or mathematically ambiguous samples are unlikely to be favored by the CVS protocol. If a sample is noisy, unclear, or semantically inconsistent, introducing the question typically leads to unstable verdict changes and often fails to satisfy the joint filtering condition $CVS_{\mathrm{Yes}}>0$ and $CVS_{\mathrm{No}}<0$. Conversely, if a sample is overly easy, its $CVS_{\mathrm{Yes}}$ tends to be large, indicating that the evaluator can validate the answer with high confidence once the question is provided; such samples are not prioritized by our low-score strategy. Therefore, CVS favors samples of moderate difficulty: they exhibit a small positive affirmation shift together with a negative rejection shift, suggesting that the question provides meaningful but non-trivial guidance for vision-language reasoning. This makes them more suitable for effective visual instruction tuning.

\section{Connection to Conditional Information Gain}

CVS is information-gain inspired, but it is not intended to be an exact estimator of the conditional mutual information $I(Q;Y\mid I)$. Estimating $I(Q;Y\mid I)$ would require access to the true data distribution and marginalization over possible questions and verdicts, which is generally infeasible for large-scale visual instruction data. Instead, CVS uses the conditional likelihood ratio
\begin{equation}
\log \frac{P(Y\mid I,Q,A)}{P(Y\mid I,A)}
\end{equation}
as a computable sample-level proxy for measuring how much the question $Q$ changes the evaluator's answer-validity judgment under the same visual context $I$.

Therefore, the goal of CVS is not to recover the absolute value of conditional mutual information, but to obtain a stable ranking signal for data selection. A positive affirmation shift indicates that the question provides additional support for the answer, while a positive rejection shift indicates possible semantic conflict. By combining these two relative shifts, CVS operationally distinguishes informative cross-modal samples from easy, noisy, or misaligned samples.

\begin{table}[htbp]
  \centering
  \caption{Main experimental results on Vision-Flan. We compare different data selection methods on Vision-Flan using 5\%, 10\%, and 15\% data subsets. The best result among subset-selection methods in each column is highlighted in bold.}
  \label{tab:full_comparison_vf}

  \begin{subtable}{\textwidth}
    \centering
    \caption{Detailed results of different methods on Vision-Flan using a 5\% data subset.}
    \label{tab:vfsubset_5}
    \resizebox{\textwidth}{!}{%
      \begin{tabular}{lcccccccccccccc}
      \toprule
      \textbf{Method} & \textbf{GQA} & \textbf{VizWiz} & \textbf{TextVQA} & \textbf{SQA-I} & \textbf{MME} & \textbf{MMB-CN} & \textbf{MMB-EN} & \textbf{AI2D} & \textbf{ChartQA} & \textbf{DocVQA} & \textbf{InfoVQA} & \textbf{MMStar} & \textbf{OCRBench} & \textbf{ARP} \\
      \midrule
        Full Data & 47.30 & 55.88 & 35.83 & 61.43 & 1270.40 & 50.26 & 55.67 & 52.14 & 15.68 & 15.80 & 15.53 & 35.32 & 26.20 & 100.00 \\
        \midrule
        Random & 42.93 & 55.79 & 36.54 & 60.34 & 1098.02 & 36.40 & 26.19 & 37.82 & 15.96 & 16.46 & 19.99 & 34.77 & 27.80 & 92.96 \\
        XMAS & 40.38 & 54.15 & 37.06 & 62.42 & 705.48 & 39.95 & 39.95 & 41.22 & 15.40 & 17.98 & 19.80 & 34.37 & 27.50 & 93.46 \\
        COINCIDE & 42.15 & 53.48 & \textbf{38.00} & 55.03 & 1000.71 & 29.12 & 41.24 & 36.40 & 16.64 & 17.89 & 17.79 & 33.51 & 28.70 & 92.26 \\
        D2Prune & 41.02 & 55.99 & 37.47 & 60.59 & 928.95 & 36.94 & 44.93 & 38.54 & 16.60 & 17.68 & 20.58 & 33.47 & 27.90 & 95.61 \\
        EL2N & \textbf{43.26} & 54.93 & 35.94 & 54.34 & 898.77 & 39.78 & 18.90 & 34.03 & 15.80 & 16.23 & 18.36 & 34.13 & 28.20 & 88.74 \\
        SemDeDup & 43.09 & 55.72 & 37.31 & 63.11 & \textbf{1115.77} & 41.32 & 48.71 & 41.16 & \textbf{17.80} & 19.02 & 19.42 & 33.09 & 27.80 & \textbf{99.45} \\
        CLIP Score & 39.24 & 54.77 & 31.80 & 57.71 & 549.88 & 38.06 & 40.38 & 37.63 & 15.92 & 16.78 & 19.69 & 33.75 & \textbf{29.20} & 89.92 \\
        SCALSELECT & 39.52 & 53.96 & 36.08 & 62.87 & 838.18 & 38.49 & 43.64 & \textbf{43.13} & 15.28 & 17.67 & 19.07 & 34.01 & 27.50 & 93.86 \\
        PRISM & 29.97 & 52.06 & 26.73 & 49.33 & 1106.50 & 35.13 & 16.67 & 29.99 & 17.00 & \textbf{19.56} & 19.35 & 35.26 & 27.90 & 86.08 \\
        \rowcolor{lightcyan}
        CVS (Ours) & 38.42 & \textbf{57.46} & 37.98 & \textbf{63.41} & 800.81 & \textbf{45.88} & \textbf{52.66} & 39.44 & 15.12 & 18.48 & \textbf{20.74} & \textbf{35.59} & 27.80 & 97.82 \\
        \bottomrule
      \end{tabular}%
    }
  \end{subtable}
  
  \vspace{10pt} 

  \begin{subtable}{\textwidth}
    \centering
    \caption{Detailed results of different methods on Vision-Flan using a 10\% data subset.}
    \label{tab:vfsubset_10}
    \resizebox{\textwidth}{!}{%
      \begin{tabular}{lcccccccccccccc}
      \toprule
      \textbf{Method} & \textbf{GQA} & \textbf{VizWiz} & \textbf{TextVQA} & \textbf{SQA-I} & \textbf{MME} & \textbf{MMB-CN} & \textbf{MMB-EN} & \textbf{AI2D} & \textbf{ChartQA} & \textbf{DocVQA} & \textbf{InfoVQA} & \textbf{MMStar} & \textbf{OCRBench} & \textbf{ARP} \\
      \midrule
        Full Data & 47.30 & 55.88 & 35.83 & 61.43 & 1270.40 & 50.26 & 55.67 & 52.14 & 15.68 & 15.80 & 15.53 & 35.32 & 26.20 & 100.00 \\
        \midrule
        Random & 43.31 & 56.02 & 36.13 & 60.92 & 1152.77 & 39.61 & 31.88 & 44.50 & \textbf{16.52} & 17.27 & 18.08 & 34.36 & 27.90 & 95.30 \\
        XMAS & 42.73 & 55.82 & 36.55 & 64.40 & 1008.41 & 45.19 & 54.38 & 43.95 & 16.20 & 18.39 & 18.19 & 34.28 & 28.50 & 99.31 \\
        COINCIDE & 43.33 & 57.38 & 38.70 & 60.83 & 1032.00 & 43.56 & 41.24 & 40.16 & 15.84 & 18.96 & 19.25 & 34.27 & 29.00 & 97.93 \\
        D2Prune & 43.47 & 56.40 & 37.81 & 63.01 & 1146.50 & \textbf{47.68} & 51.80 & 43.01 & 16.44 & 17.38 & 18.78 & 34.73 & 29.00 & 100.49 \\
        EL2N & \textbf{44.08} & 56.99 & 38.10 & 47.99 & 1088.62 & 39.52 & 31.70 & 29.92 & 15.64 & 18.78 & 18.78 & 34.94 & 28.30 & 92.67 \\
        SemDeDup & 43.93 & 56.39 & 37.20 & 58.65 & \textbf{1236.25} & 47.51 & 45.96 & 39.31 & 15.72 & 18.89 & 17.58 & 36.00 & 28.50 & 98.97 \\
        CLIP Score & 41.02 & 53.63 & 38.11 & 59.84 & 725.09 & 34.28 & 30.50 & 35.56 & 16.36 & 20.29 & 22.72 & 34.90 & \textbf{29.10} & 94.13 \\
        SCALSELECT & 42.73 & 56.26 & 37.68 & 60.88 & 1101.46 & 44.42 & \textbf{54.81} & 44.49 & 14.84 & 18.29 & 16.71 & 33.56 & 27.80 & 97.94 \\
        PRISM & 35.21 & 48.80 & 37.87 & 61.97 & 751.61 & 38.49 & 44.16 & \textbf{44.88} & 16.44 & \textbf{23.76} & \textbf{23.37} & 35.92 & 28.80 & 98.99 \\
        \rowcolor{lightcyan}
        CVS (Ours) & 39.49 & \textbf{57.66} & \textbf{38.99} & \textbf{66.78} & 930.81 & 46.13 & 54.30 & 43.98 & 16.24 & 21.18 & 21.72 & \textbf{37.70} & 28.70 & \textbf{103.45} \\
        \bottomrule
      \end{tabular}%
    }
  \end{subtable}

  \vspace{10pt} 

  \begin{subtable}{\textwidth}
    \centering
    \caption{Detailed results of different methods on Vision-Flan using a 15\% data subset.}
    \label{tab:vfsubset_15}
    \resizebox{\textwidth}{!}{%
      \begin{tabular}{lcccccccccccccc}
      \toprule
      \textbf{Method} & \textbf{GQA} & \textbf{VizWiz} & \textbf{TextVQA} & \textbf{SQA-I} & \textbf{MME} & \textbf{MMB-CN} & \textbf{MMB-EN} & \textbf{AI2D} & \textbf{ChartQA} & \textbf{DocVQA} & \textbf{InfoVQA} & \textbf{MMStar} & \textbf{OCRBench} & \textbf{ARP} \\
      \midrule
        Full Data & 47.30 & 55.88 & 35.83 & 61.43 & 1270.40 & 50.26 & 55.67 & 52.14 & 15.68 & 15.80 & 15.53 & 35.32 & 26.20 & 100.00 \\
        \midrule
        Random & 43.19 & 56.33 & 35.41 & 59.59 & 1230.04 & 42.44 & 38.06 & 41.06 & 15.52 & 16.26 & 16.26 & 32.27 & 27.60 & 93.82 \\
        XMAS & 44.68 & 56.52 & 37.19 & 61.18 & 1012.67 & 40.72 & 46.99 & 44.62 & 15.84 & 17.83 & 17.88 & 34.95 & 27.50 & 97.13 \\
        COINCIDE & 44.28 & 56.44 & 37.52 & 55.73 & 1221.20 & 45.96 & 39.26 & 37.50 & 17.24 & 19.03 & 18.32 & 35.59 & 28.80 & 98.39 \\
        D2Prune & 43.96 & 57.16 & 38.40 & 61.63 & 1059.64 & 44.67 & 49.48 & 44.66 & 15.80 & 17.62 & 18.60 & 34.13 & 28.30 & 98.94 \\
        EL2N & 45.81 & 57.45 & 38.52 & 53.69 & 1105.20 & 46.22 & 39.18 & 43.39 & 14.68 & 19.06 & 17.35 & 34.56 & 28.40 & 96.87 \\
        SemDeDup & \textbf{50.00} & 51.57 & 37.30 & 59.69 & \textbf{1247.46} & 43.90 & 45.45 & 43.13 & 17.04 & 18.66 & 18.93 & 35.24 & 28.00 & 100.35 \\
        CLIP Score & 41.33 & 55.10 & 38.44 & \textbf{65.25} & 640.09 & 39.60 & 41.67 & \textbf{48.41} & 16.68 & 20.19 & 21.96 & 37.57 & \textbf{29.30} & 99.24 \\
        SCALSELECT & 43.15 & 57.90 & 38.47 & 57.56 & 1062.93 & 46.91 & 43.99 & 36.40 & 15.96 & 18.82 & 16.69 & 35.53 & 27.70 & 96.65 \\
        PRISM & 39.25 & 52.24 & 38.62 & 61.23 & 770.61 & 41.67 & 41.07 & 43.78 & \textbf{17.96} & 21.12 & 21.00 & 36.22 & 29.10 & 98.64 \\
        \rowcolor{lightcyan}
        CVS (Ours) & 43.46 & \textbf{57.93} & \textbf{40.57} & 63.46 & 1123.40 & \textbf{47.17} & \textbf{51.03} & 42.23 & 16.00 & \textbf{21.36} & \textbf{21.98} & \textbf{37.64} & 29.00 & \textbf{104.85} \\
        \bottomrule
      \end{tabular}%
    }
  \end{subtable}

\end{table}


\begin{table*}[htbp]
  \centering
  \caption{Main experimental results on The Cauldron. We compare different data selection methods on The Cauldron using 10\%, 20\%, 30\%, and 50\% data subsets. The best result in each column is highlighted in bold.}
  \label{tab:full_comparison_cr}

  \begin{subtable}{\textwidth}
    \centering
    \caption{Detailed results of different methods on The Cauldron using a 10\% data subset.}
    \label{tab:crsubset_10}
    \resizebox{\textwidth}{!}{%
      \begin{tabular}{lcccccccccccccc}
      \toprule
      \textbf{Method} & \textbf{GQA} & \textbf{VizWiz} & \textbf{TextVQA} & \textbf{SQA-I} & \textbf{MME} & \textbf{MMB-CN} & \textbf{MMB-EN} & \textbf{AI2D} & \textbf{ChartQA} & \textbf{DocVQA} & \textbf{InfoVQA} & \textbf{MMStar} & \textbf{OCRBench} & \textbf{ARP} \\
      \midrule
        Full Data & 46.76 & 58.16 & 49.13 & 78.33 & 1333.04 & 57.73 & 65.72 & 64.57 & 12.80 & 27.52 & 15.88 & 40.38 & 35.10 & 100.00 \\
        \midrule
        Random & 44.10 & \textbf{48.73} & 42.82 & 66.23 & 1038.67 & 45.73 & 57.19 & 50.91 & 10.24 & 20.55 & 14.03 & 34.08 & 30.10 & 83.54 \\
        XMAS & \textbf{44.56} & 41.38 & 44.10 & \textbf{69.11} & 1104.39 & 48.37 & 56.27 & 52.59 & 10.32 & 20.62 & 14.14 & 34.26 & 30.40 & \textbf{84.17} \\
        COINCIDE & 43.85 & 45.39 & 43.40 & 67.17 & \textbf{1118.15} & 48.48 & 54.60 & 50.39 & 10.40 & 20.31 & 14.39 & 34.15 & 30.20 & 83.94 \\
        D2Prune & 43.06 & 43.29 & 43.35 & 68.91 & 1009.67 & 47.62 & 56.76 & 52.59 & 10.12 & 19.68 & 13.89 & 34.60 & 28.80 & 82.66 \\
        EL2N & 43.55 & 41.20 & 43.94 & 68.95 & 1072.84 & 48.39 & 55.03 & 50.26 & \textbf{10.68} & 20.78 & \textbf{14.80} & \textbf{35.10} & 30.50 & 84.10 \\
        SemDeDup & 43.02 & 45.04 & 43.24 & 69.00 & 1003.83 & \textbf{48.65} & \textbf{57.96} & 51.33 & 9.40 & 20.43 & 14.48 & 33.87 & 30.10 & 83.18 \\
        CLIP Score & 44.08 & 38.73 & \textbf{44.13} & 67.42 & 1072.95 & 48.05 & 55.29 & \textbf{56.56} & 9.08 & 19.90 & 13.67 & 34.82 & 30.40 & 82.64 \\
        \rowcolor{lightcyan}
        CVS (Ours) & 43.58 & 40.58 & 43.90 & 67.64 & 1000.09 & 48.25 & 56.88 & 53.79 & 9.64 & \textbf{21.15} & 14.45 & 34.77 & \textbf{31.40} & 83.53 \\
        \bottomrule
      \end{tabular}%
    }
  \end{subtable}
  
  \vspace{10pt}

  \begin{subtable}{\textwidth}
    \centering
    \caption{Detailed results of different methods on The Cauldron using a 20\% data subset.}
    \label{tab:crsubset_20}
    \resizebox{\textwidth}{!}{%
      \begin{tabular}{lcccccccccccccc}
      \toprule
      \textbf{Method} & \textbf{GQA} & \textbf{VizWiz} & \textbf{TextVQA} & \textbf{SQA-I} & \textbf{MME} & \textbf{MMB-CN} & \textbf{MMB-EN} & \textbf{AI2D} & \textbf{ChartQA} & \textbf{DocVQA} & \textbf{InfoVQA} & \textbf{MMStar} & \textbf{OCRBench} & \textbf{ARP} \\
      \midrule
        Full Data & 46.76 & 58.16 & 49.13 & 78.33 & 1333.04 & 57.73 & 65.72 & 64.57 & 12.80 & 27.52 & 15.88 & 40.38 & 35.10 & 100.00 \\
        \midrule
        Random & 44.50 & 49.20 & 44.78 & 68.27 & 1116.78 & 50.80 & 58.47 & 56.09 & 10.64 & 22.23 & 15.23 & 35.23 & \textbf{31.90} & \textbf{87.96} \\
        XMAS & 44.20 & 42.83 & \textbf{45.80} & 71.24 & 1086.13 & 49.83 & 57.30 & 55.18 & 10.60 & 22.77 & \textbf{15.58} & 35.14 & 30.60 & 86.97 \\
        COINCIDE & \textbf{45.04} & 48.11 & 45.05 & 70.96 & \textbf{1152.90} & 50.03 & 56.92 & 53.53 & 9.96 & 22.04 & 14.96 & 33.45 & 30.60 & 86.63 \\
        D2Prune & 43.83 & 43.58 & 44.33 & \textbf{71.63} & 1131.72 & 51.32 & 57.70 & 55.21 & \textbf{11.48} & 21.83 & 14.44 & 31.78 & 30.40 & 86.35 \\
        EL2N & 43.67 & 43.36 & 44.45 & 70.49 & 1101.89 & 50.55 & 57.10 & 55.86 & 10.76 & 22.27 & 14.62 & 33.73 & 30.90 & 86.20 \\
        SemDeDup & 44.11 & \textbf{49.40} & 44.89 & 70.49 & 1041.51 & 49.68 & 56.67 & 55.11 & 10.36 & 22.44 & 14.87 & 34.44 & 31.50 & 86.73 \\
        CLIP Score & 44.74 & 37.79 & 45.11 & 71.28 & 1144.76 & \textbf{51.74} & \textbf{59.50} & \textbf{59.71} & 10.00 & 21.90 & 14.89 & \textbf{37.07} & 30.70 & 87.12 \\
        \rowcolor{lightcyan}
        CVS (Ours) & 44.57 & 45.84 & 45.03 & 70.51 & 1035.29 & 50.23 & 56.19 & 57.09 & 10.56 & \textbf{22.92} & 15.52 & 34.64 & 31.00 & 87.09 \\
        \bottomrule
      \end{tabular}%
    }
  \end{subtable}

  \vspace{10pt} 

  \begin{subtable}{\textwidth}
    \centering
    \caption{Detailed results of different methods on The Cauldron using a 30\% data subset.}
    \label{tab:crsubset_30}
    \resizebox{\textwidth}{!}{%
      \begin{tabular}{lcccccccccccccc}
      \toprule
      \textbf{Method} & \textbf{GQA} & \textbf{VizWiz} & \textbf{TextVQA} & \textbf{SQA-I} & \textbf{MME} & \textbf{MMB-CN} & \textbf{MMB-EN} & \textbf{AI2D} & \textbf{ChartQA} & \textbf{DocVQA} & \textbf{InfoVQA} & \textbf{MMStar} & \textbf{OCRBench} & \textbf{ARP} \\
      \midrule
        Full Data & 46.76 & 58.16 & 49.13 & 78.33 & 1333.04 & 57.73 & 65.72 & 64.57 & 12.80 & 27.52 & 15.88 & 40.38 & 35.10 & 100.00 \\
        \midrule
        Random & 45.15 & 44.57 & 45.66 & 72.23 & 1188.38 & 51.49 & 58.22 & 55.61 & \textbf{11.08} & 23.07 & 15.62 & 34.45 & 32.00 & 88.97 \\
        XMAS & 44.89 & 47.14 & \textbf{46.60} & 72.78 & 1117.63 & \textbf{52.92} & \textbf{61.51} & 58.52 & 10.92 & 24.02 & 15.06 & 35.76 & 32.50 & 90.24 \\
        COINCIDE & 45.11 & 50.12 & 45.29 & \textbf{73.61} & \textbf{1235.28} & 52.69 & 60.36 & 57.93 & 10.84 & \textbf{25.39} & 15.55 & 34.35 & 31.80 & \textbf{91.14} \\
        D2Prune & 44.18 & 44.52 & 45.83 & 73.07 & 1126.04 & 51.75 & 60.28 & 57.78 & 10.00 & 22.45 & 14.86 & \textbf{35.90} & 32.20 & 88.22 \\
        EL2N & \textbf{45.83} & \textbf{50.83} & 45.16 & 71.73 & 1213.79 & 51.91 & 60.45 & 57.64 & 10.72 & 23.35 & 14.77 & 33.73 & 30.90 & 89.56 \\
        SemDeDup & 44.34 & 42.59 & 44.98 & 72.08 & 1164.76 & 52.34 & 60.62 & 56.93 & \textbf{11.08} & 23.33 & 15.04 & 35.80 & 31.70 & 88.85 \\
        CLIP Score & 45.31 & 39.77 & 45.85 & 71.38 & 1235.07 & 52.77 & 61.39 & \textbf{60.52} & 10.56 & 22.86 & 15.09 & 35.68 & 32.30 & 89.38 \\
        \rowcolor{lightcyan}
        CVS (Ours) & 44.92 & 50.73 & 45.14 & 71.36 & 1095.65 & 49.89 & 59.97 & 59.11 & 10.64 & 23.48 & \textbf{15.78} & 35.41 & \textbf{32.70} & 89.72 \\
        \bottomrule
      \end{tabular}%
    }
  \end{subtable}

  \vspace{10pt} 

  \begin{subtable}{\textwidth}
    \centering
    \caption{Detailed results of different methods on The Cauldron using a 50\% data subset.}
    \label{tab:crsubset_50}
    \resizebox{\textwidth}{!}{%
      \begin{tabular}{lcccccccccccccc}
      \toprule
      \textbf{Method} & \textbf{GQA} & \textbf{VizWiz} & \textbf{TextVQA} & \textbf{SQA-I} & \textbf{MME} & \textbf{MMB-CN} & \textbf{MMB-EN} & \textbf{AI2D} & \textbf{ChartQA} & \textbf{DocVQA} & \textbf{InfoVQA} & \textbf{MMStar} & \textbf{OCRBench} & \textbf{ARP} \\
      \midrule
        Full Data & 46.76 & 58.16 & 49.13 & 78.33 & 1333.04 & 57.73 & 65.72 & 64.57 & 12.80 & 27.52 & 15.88 & 40.38 & 35.10 & 100.00 \\
        \midrule
        Random & 45.38 & 49.75 & 46.60 & 73.97 & 1136.19 & 53.47 & 61.40 & 60.50 & 11.32 & 24.69 & 15.06 & 36.25 & 32.70 & 91.75 \\
        XMAS & 45.44 & 44.92 & \textbf{46.72} & 75.11 & 1223.29 & 54.12 & 61.94 & 59.23 & 11.56 & \textbf{25.24} & 15.33 & 35.20 & 33.10 & 92.07 \\
        COINCIDE & \textbf{46.55} & 51.45 & 46.60 & 75.25 & 1236.37 & \textbf{55.09} & 62.51 & 60.33 & 11.32 & 24.14 & 14.92 & 35.91 & 33.60 & 93.11 \\
        D2Prune & 45.16 & \textbf{54.69} & 45.96 & 73.57 & 1258.27 & 53.63 & 62.16 & 59.26 & \textbf{12.04} & 24.70 & 15.10 & 35.99 & 33.70 & \textbf{93.52} \\
        EL2N & 45.21 & 47.63 & 46.30 & 74.41 & 1222.04 & 53.46 & 61.65 & 59.16 & 10.52 & 24.69 & 15.19 & 35.10 & 33.70 & 91.39 \\
        SemDeDup & 45.39 & 47.50 & 46.68 & \textbf{75.26} & \textbf{1281.12} & 54.92 & 62.25 & 59.36 & 11.32 & 24.67 & \textbf{15.47} & 36.47 & 32.50 & 92.78 \\
        CLIP Score & 45.98 & 45.73 & 46.46 & 73.07 & 1215.97 & 53.20 & \textbf{63.11} & 60.81 & 11.96 & 24.27 & 14.83 & 37.22 & 32.10 & 92.07 \\
        \rowcolor{lightcyan}
        CVS (Ours) & 44.99 & 45.33 & 45.86 & 75.22 & 1181.08 & 54.10 & 62.89 & \textbf{61.69} & 10.72 & 24.88 & 14.95 & \textbf{38.25} & \textbf{34.40} & 92.16 \\
        \bottomrule
      \end{tabular}%
    }
  \end{subtable}

\end{table*}


\begin{table*}[htbp]
  \centering
  \caption{Detailed results on Vision-Flan for the effect of CVS score ranges. We evaluate different CVS score selection strategies (Low / High / No) on the Vision-Flan using data subsets of 5\%, 10\%, and 15\%. The best result in each column is highlighted in bold.}
  \label{tab:full_comparison_lhn}

  \begin{subtable}{\textwidth}
    \centering
    \caption{Detailed results of different CVS score selection strategies on Vision-Flan using a 5\% data subset.}
    \label{tab:lhnsubset_5}
    \resizebox{\textwidth}{!}{%
      \begin{tabular}{lcccccccccccccc}
      \toprule
      \textbf{Method} & \textbf{GQA} & \textbf{VizWiz} & \textbf{TextVQA} & \textbf{SQA-I} & \textbf{MME} & \textbf{MMB-CN} & \textbf{MMB-EN} & \textbf{AI2D} & \textbf{ChartQA} & \textbf{DocVQA} & \textbf{InfoVQA} & \textbf{MMStar} & \textbf{OCRBench} & \textbf{ARP} \\
      \midrule
        Full Data & 47.30 & 55.88 & 35.83 & 61.43 & 1270.40 & 50.26 & 55.67 & 52.14 & 15.68 & 15.80 & 15.53 & 35.32 & 26.20 & 100.00 \\
        \midrule
        Random & \textbf{42.93} & 55.79 & 36.54 & 60.34 & \textbf{1098.02} & 36.40 & 26.19 & 37.82 & \textbf{15.96} & 16.46 & 19.99 & 34.77 & 27.80 & 92.96 \\
        No & 24.31 & 53.87 & 29.59 & 53.20 & 564.60 & 31.70 & 38.66 & 27.14 & 14.80 & 13.43 & 15.22 & 35.02 & \textbf{27.90} & 79.16 \\
        High & 35.07 & 51.09 & 30.75 & 62.27 & 488.55 & 39.69 & 44.16 & \textbf{45.95} & 15.12 & 14.55 & 19.77 & 31.38 & 23.80 & 87.16 \\
        Low & 38.42 & \textbf{57.46} & \textbf{37.98} & \textbf{63.41} & 800.81 & \textbf{45.88} & \textbf{52.66} & 39.44 & 15.12 & \textbf{18.48} & \textbf{20.74} & \textbf{35.59} & 27.80 & \textbf{97.82} \\

        \bottomrule
      \end{tabular}%
    }
  \end{subtable}
  
  \vspace{10pt}

  \begin{subtable}{\textwidth}
    \centering
    \caption{Detailed results of different CVS score selection strategies on Vision-Flan using a 10\% data subset.}
    \label{tab:lhnsubset_10}
    \resizebox{\textwidth}{!}{%
      \begin{tabular}{lcccccccccccccc}
      \toprule
      \textbf{Method} & \textbf{GQA} & \textbf{VizWiz} & \textbf{TextVQA} & \textbf{SQA-I} & \textbf{MME} & \textbf{MMB-CN} & \textbf{MMB-EN} & \textbf{AI2D} & \textbf{ChartQA} & \textbf{DocVQA} & \textbf{InfoVQA} & \textbf{MMStar} & \textbf{OCRBench} & \textbf{ARP} \\
      \midrule
        Full Data & 47.30 & 55.88 & 35.83 & 61.43 & 1270.40 & 50.26 & 55.67 & 52.14 & 15.68 & 15.80 & 15.53 & 35.32 & 26.20 & 100.00 \\
        \midrule
        Random & \textbf{43.31} & 56.02 & 36.13 & 60.92 & \textbf{1152.77} & 39.61 & 31.88 & 44.50 & \textbf{16.52} & 17.27 & 18.08 & 34.36 & 27.90 & 95.30 \\
        No & 34.72 & 54.65 & 36.96 & 48.88 & 870.97 & 32.90 & 40.89 & 22.63 & 15.52 & 17.80 & 20.56 & 37.18 & 28.60 & 89.48 \\
        High & 38.09 & 55.15 & 33.52 & 59.79 & 533.81 & 45.79 & 36.00 & \textbf{44.66} & 15.44 & 15.81 & 18.22 & 32.50 & 25.20 & 89.05 \\
        Low & 39.49 & \textbf{57.66} & \textbf{38.99} & \textbf{66.78} & 930.81 & \textbf{46.13} & \textbf{54.30} & 43.98 & 16.24 & \textbf{21.18} & \textbf{21.72} & \textbf{37.70} & \textbf{28.70} & \textbf{103.45} \\
        \bottomrule
      \end{tabular}%
    }
  \end{subtable}

  \vspace{10pt} 

  \begin{subtable}{\textwidth}
    \centering
    \caption{Detailed results of different CVS score selection strategies on Vision-Flan using a 15\% data subset.}
    \label{tab:lhnsubset_15}
    \resizebox{\textwidth}{!}{%
      \begin{tabular}{lcccccccccccccc}
      \toprule
      \textbf{Method} & \textbf{GQA} & \textbf{VizWiz} & \textbf{TextVQA} & \textbf{SQA-I} & \textbf{MME} & \textbf{MMB-CN} & \textbf{MMB-EN} & \textbf{AI2D} & \textbf{ChartQA} & \textbf{DocVQA} & \textbf{InfoVQA} & \textbf{MMStar} & \textbf{OCRBench} & \textbf{ARP} \\
      \midrule
        Full Data & 47.30 & 55.88 & 35.83 & 61.43 & 1270.40 & 50.26 & 55.67 & 52.14 & 15.68 & 15.80 & 15.53 & 35.32 & 26.20 & 100.00 \\
        \midrule
        Random & 43.19 & 56.33 & 35.41 & 59.59 & \textbf{1230.04} & 42.44 & 38.06 & 41.06 & 15.52 & 16.26 & 16.26 & 32.27 & 27.60 & 93.82 \\
        No & 36.61 & 55.70 & 37.60 & 62.42 & 668.22 & 40.21 & 39.69 & 39.25 & \textbf{16.88} & 16.41 & 18.21 & 36.19 & \textbf{29.30} & 92.76 \\
        High & 40.32 & 54.78 & 33.47 & 63.26 & 830.61 & 45.70 & \textbf{53.01} & \textbf{53.69} & 15.44 & 16.26 & 17.12 & 35.81 & 26.40 & 96.00 \\
        Low & \textbf{43.46} & \textbf{57.93} & \textbf{40.57} & \textbf{63.46} & 1123.40 & \textbf{47.17} & 51.03 & 42.23 & 16.00 & \textbf{21.36} & \textbf{21.98} & \textbf{37.64} & 29.00 & \textbf{104.85} \\
        \bottomrule
      \end{tabular}%
    }
  \end{subtable}

\end{table*}

\begin{table*}[htbp]
  \centering
  \caption{Detailed results on Vision-Flan under different evaluator architectures.
We compare different evaluator architectures on Vision-Flan using data subsets of 5\%, 10\%, and 15\%. Qwen, InternVL, and LLaVA-OV correspond to Qwen2.5-VL-7B-Instruct, InternVL3-8B, and LLaVA-OneVision-1.5-4B-Instruct, respectively. The best result among subset-selection methods in each column is highlighted in bold.}
  \label{tab:full_comparison_arch}

  \begin{subtable}{\textwidth}
    \centering
    \caption{Detailed results of different evaluator architectures on Vision-Flan using a 5\% data subset.}
    \label{tab:archsubset_5}
    \resizebox{\textwidth}{!}{%
      \begin{tabular}{lcccccccccccccc}
      \toprule
      \textbf{Method} & \textbf{GQA} & \textbf{VizWiz} & \textbf{TextVQA} & \textbf{SQA-I} & \textbf{MME} & \textbf{MMB-CN} & \textbf{MMB-EN} & \textbf{AI2D} & \textbf{ChartQA} & \textbf{DocVQA} & \textbf{InfoVQA} & \textbf{MMStar} & \textbf{OCRBench} & \textbf{ARP} \\
      \midrule
        Full Data & 47.30 & 55.88 & 35.83 & 61.43 & 1270.40 & 50.26 & 55.67 & 52.14 & 15.68 & 15.80 & 15.53 & 35.32 & 26.20 & 100.00 \\
        \midrule
        Random & \textbf{42.93} & 55.79 & 36.54 & 60.34 & \textbf{1098.02} & 36.40 & 26.19 & 37.82 & \textbf{15.96} & 16.46 & 19.99 & 34.77 & 27.80 & 92.96 \\
        CVS(Qwen) & 38.42 & 57.46 & \textbf{37.98} & 63.41 & 800.81 & 45.88 & \textbf{52.66} & 39.44 & 15.12 & \textbf{18.48} & \textbf{20.74} & 35.59 & 27.80 & \textbf{97.82} \\
        CVS(InternVL) & 36.93 & \textbf{57.97} & 34.42 & 60.09 & 839.81 & \textbf{46.48} & 43.73 & 39.67 & 14.56 & 16.34 & 18.24 & \textbf{38.27} & \textbf{28.70} & 93.89 \\
        CVS(LLaVA-OV) & 39.01 & 54.20 & 34.99 & \textbf{64.51} & 1084.98 & 35.05 & 45.27 & \textbf{47.96} & 14.76 & 17.50 & 20.58 & 31.17 & 27.20 & 95.39 \\
        \bottomrule
      \end{tabular}%
    }
  \end{subtable}
  
  \vspace{10pt} 

  \begin{subtable}{\textwidth}
    \centering
    \caption{Detailed results of different evaluator architectures on Vision-Flan using a 10\% data subset.}
    \label{tab:arsubset_10}
    \resizebox{\textwidth}{!}{%
      \begin{tabular}{lcccccccccccccc}
      \toprule
      \textbf{Method} & \textbf{GQA} & \textbf{VizWiz} & \textbf{TextVQA} & \textbf{SQA-I} & \textbf{MME} & \textbf{MMB-CN} & \textbf{MMB-EN} & \textbf{AI2D} & \textbf{ChartQA} & \textbf{DocVQA} & \textbf{InfoVQA} & \textbf{MMStar} & \textbf{OCRBench} & \textbf{ARP} \\
      \midrule
        Full Data & 47.30 & 55.88 & 35.83 & 61.43 & 1270.40 & 50.26 & 55.67 & 52.14 & 15.68 & 15.80 & 15.53 & 35.32 & 26.20 & 100.00 \\
        \midrule
        Random & \textbf{43.31} & 56.02 & 36.13 & 60.92 & \textbf{1152.77} & 39.61 & 31.88 & \textbf{44.50} & 16.52 & 17.27 & 18.08 & 34.36 & 27.90 & 95.30 \\
        CVS(Qwen) & 39.49 & \textbf{57.66} & 38.99 & \textbf{66.78} & 930.81 & 46.13 & \textbf{54.30} & 43.98 & 16.24 & 21.18 & \textbf{21.72} & 37.70 & 28.70 & \textbf{103.45} \\
        CVS(InternVL) & 36.71 & 54.36 & \textbf{39.91} & 61.82 & 631.84 & \textbf{48.72} & 43.38 & 40.16 & \textbf{16.56} & 18.75 & 19.06 & \textbf{39.08} & 29.00 & 96.68 \\
        CVS(LLaVA-OV) & 37.02 & 54.21 & 39.65 & 60.34 & 871.82 & 40.71 & 34.02 & 39.12 & 15.36 & \textbf{21.52} & 21.11 & 35.31 & \textbf{30.30} & 96.59 \\
        \bottomrule
      \end{tabular}%
    }
  \end{subtable}

  \vspace{10pt} 

  \begin{subtable}{\textwidth}
    \centering
    \caption{Detailed results of different evaluator architectures on Vision-Flan using a 15\% data subset.}
    \label{tab:arsubset_15}
    \resizebox{\textwidth}{!}{%
      \begin{tabular}{lcccccccccccccc}
      \toprule
      \textbf{Method} & \textbf{GQA} & \textbf{VizWiz} & \textbf{TextVQA} & \textbf{SQA-I} & \textbf{MME} & \textbf{MMB-CN} & \textbf{MMB-EN} & \textbf{AI2D} & \textbf{ChartQA} & \textbf{DocVQA} & \textbf{InfoVQA} & \textbf{MMStar} & \textbf{OCRBench} & \textbf{ARP} \\
      \midrule
        Full Data & 47.30 & 55.88 & 35.83 & 61.43 & 1270.40 & 50.26 & 55.67 & 52.14 & 15.68 & 15.80 & 15.53 & 35.32 & 26.20 & 100.00 \\
        \midrule
        Random & 43.19 & 56.33 & 35.41 & 59.59 & \textbf{1230.04} & 42.44 & 38.06 & 41.06 & 15.52 & 16.26 & 16.26 & 32.27 & 27.60 & 93.82 \\
        CVS(Qwen) & \textbf{43.46} & \textbf{57.93} & 40.57 & \textbf{63.46} & 1123.40 & 47.17 & \textbf{51.03} & \textbf{42.23} & 16.00 & 21.36 & \textbf{21.98} & \textbf{37.64} & 29.00 & \textbf{104.85} \\
        CVS(InternVL) & 35.20 & 57.04 & 39.61 & 61.58 & 550.58 & \textbf{49.91} & 50.00 & 41.48 & \textbf{17.00} & 20.44 & 19.65 & 37.32 & 29.10 & 98.49 \\
        CVS(LLaVA-OV) & 39.05 & 57.20 & \textbf{40.87} & 60.78 & 524.28 & 45.36 & 44.50 & 41.22 & 15.60 & \textbf{21.73} & 21.52 & 36.30 & \textbf{30.00} & 98.56 \\
        \bottomrule
      \end{tabular}%
    }
  \end{subtable}

\end{table*}

\begin{table*}[htbp]
  \centering
  \caption{Detailed results under different evaluator scales on Vision-Flan. We compare evaluators of different scales within the Qwen2.5-VL family, where CVS (7B) and CVS (3B) denote using Qwen2.5-VL-7B-Instruct and Qwen2.5-VL-3B-Instruct as evaluators, respectively. Experiments are conducted using 5\%, 10\%, and 15\% data subsets. The best result in each column is highlighted in bold.}
  \label{tab:as_full_comparison}

  \begin{subtable}{\textwidth}
    \centering
    \caption{Detailed results under different evaluator scales on Vision-Flan using a 5\% data subset.}
    \label{tab:assubset_5}
    \resizebox{\textwidth}{!}{%
      \begin{tabular}{lcccccccccccccc}
      \toprule
      \textbf{Method} & \textbf{GQA} & \textbf{VizWiz} & \textbf{TextVQA} & \textbf{SQA-I} & \textbf{MME} & \textbf{MMB-CN} & \textbf{MMB-EN} & \textbf{AI2D} & \textbf{ChartQA} & \textbf{DocVQA} & \textbf{InfoVQA} & \textbf{MMStar} & \textbf{OCRBench} & \textbf{ARP} \\
      \midrule
        Full Data & 47.30 & 55.88 & 35.83 & 61.43 & 1270.40 & 50.26 & 55.67 & 52.14 & 15.68 & 15.80 & 15.53 & 35.32 & 26.20 & 100.00 \\
        \midrule
        CVS (7B) & 38.42 & \textbf{57.46} & \textbf{37.98} & \textbf{63.41} & \textbf{800.81} & 45.88 & \textbf{52.66} & \textbf{39.44} & 15.12 & \textbf{18.48} & \textbf{20.74} & 35.59 & 27.80 & \textbf{97.82} \\
        CVS (3B) & \textbf{39.17} & 53.22 & 34.89 & 62.33 & 500.00 & \textbf{46.22} & 43.56 & 39.18 & \textbf{15.20} & 15.48 & 17.74 & \textbf{37.50} & \textbf{28.60} & 91.24 \\

        \bottomrule
      \end{tabular}%
    }
  \end{subtable}
  
  \vspace{10pt}

  \begin{subtable}{\textwidth}
    \centering
    \caption{Detailed results under different evaluator scales on Vision-Flan using a 10\% data subset.}
    \label{tab:assubset_10}
    \resizebox{\textwidth}{!}{%
      \begin{tabular}{lcccccccccccccc}
      \toprule
      \textbf{Method} & \textbf{GQA} & \textbf{VizWiz} & \textbf{TextVQA} & \textbf{SQA-I} & \textbf{MME} & \textbf{MMB-CN} & \textbf{MMB-EN} & \textbf{AI2D} & \textbf{ChartQA} & \textbf{DocVQA} & \textbf{InfoVQA} & \textbf{MMStar} & \textbf{OCRBench} & \textbf{ARP} \\
      \midrule
        Full Data & 47.30 & 55.88 & 35.83 & 61.43 & 1270.40 & 50.26 & 55.67 & 52.14 & 15.68 & 15.80 & 15.53 & 35.32 & 26.20 & 100.00 \\
        \midrule
        CVS (7B) & 39.49 & \textbf{57.66} & 38.99 & \textbf{66.78} & \textbf{930.81} & \textbf{46.13} & \textbf{54.30} & \textbf{43.98} & 16.24 & \textbf{21.18} & \textbf{21.72} & 37.70 & \textbf{28.70} & \textbf{103.45} \\
        CVS (3B) & \textbf{40.13} & 53.27 & \textbf{39.65} & 63.16 & 513.92 & 43.47 & 50.43 & 41.39 & \textbf{16.84} & 18.15 & 18.28 & \textbf{39.60} & 28.40 & 96.24 \\
        \bottomrule
      \end{tabular}%
    }
  \end{subtable}

  \vspace{10pt} 

  \begin{subtable}{\textwidth}
    \centering
    \caption{Detailed results under different evaluator scales on Vision-Flan using a 15\% data subset.}
    \label{tab:assubset_15}
    \resizebox{\textwidth}{!}{%
      \begin{tabular}{lcccccccccccccc}
      \toprule
      \textbf{Method} & \textbf{GQA} & \textbf{VizWiz} & \textbf{TextVQA} & \textbf{SQA-I} & \textbf{MME} & \textbf{MMB-CN} & \textbf{MMB-EN} & \textbf{AI2D} & \textbf{ChartQA} & \textbf{DocVQA} & \textbf{InfoVQA} & \textbf{MMStar} & \textbf{OCRBench} & \textbf{ARP} \\
      \midrule
        Full Data & 47.30 & 55.88 & 35.83 & 61.43 & 1270.40 & 50.26 & 55.67 & 52.14 & 15.68 & 15.80 & 15.53 & 35.32 & 26.20 & 100.00 \\
        \midrule
        CVS (7B) & \textbf{43.46} & \textbf{57.93} & \textbf{40.57} & \textbf{63.46} & \textbf{1123.40} & \textbf{47.17} & \textbf{51.03} & \textbf{42.23} & 16.00 & \textbf{21.36} & \textbf{21.98} & \textbf{37.64} & \textbf{29.00} & \textbf{104.85} \\
        CVS (3B) & 42.61 & 55.96 & 38.41 & 62.77 & 1104.09 & 41.93 & 49.31 & 41.00 & \textbf{16.36} & 17.68 & 17.88 & 36.35 & 28.60 & 98.51 \\
        \bottomrule
      \end{tabular}%
    }
  \end{subtable}

\end{table*}

\begin{table*}[htbp]
  \centering
  \caption{Detailed results on Vision-Flan for robustness to the target model (Qwen2-VL-2B). With the evaluator fixed to Qwen2-VL-7B-Instruct, we replace the target model from LLaVA-1.5-7B to Qwen2-VL-2B to evaluate the robustness of CVS under target model changes. Experiments are conducted using data subsets of 5\%, 10\%, and 15\%. The best result in each column is highlighted in bold.}
  \label{tab:tm_full_comparison}

  \begin{subtable}{\textwidth}
    \centering
    \caption{Detailed results of CVS robustness to target model (Qwen2-VL-2B) changes using a 5\% data subset.}
    \label{tab:tmsubset_5}
    \resizebox{\textwidth}{!}{%
      \begin{tabular}{lcccccccccccccc}
      \toprule
      \textbf{Method} & \textbf{GQA} & \textbf{VizWiz} & \textbf{TextVQA} & \textbf{SQA-I} & \textbf{MME} & \textbf{MMB-CN} & \textbf{MMB-EN} & \textbf{AI2D} & \textbf{ChartQA} & \textbf{DocVQA} & \textbf{InfoVQA} & \textbf{MMStar} & \textbf{OCRBench} & \textbf{ARP} \\
      \midrule
        Full Data & 51.84 & 34.21 & 72.03 & 74.96 & 1294.90 & 61.68 & 68.99 & 68.94 & 52.00 & 81.86 & 50.67 & 42.58 & 74.80 & 100.00 \\
        \midrule
        Random & \textbf{52.40} & 40.72 & \textbf{71.88} & \textbf{75.75} & \textbf{1430.27} & 53.35 & 58.85 & \textbf{67.13} & 48.20 & 76.36 & 49.41 & \textbf{40.58} & 74.10 & 98.34 \\
        \rowcolor{lightcyan}
        CVS & 48.86 & \textbf{49.06} & 71.50 & 74.96 & 1269.53 & \textbf{55.33} & \textbf{61.25} & 67.10 & \textbf{49.12} & \textbf{79.16} & \textbf{51.58} & 39.20 & \textbf{74.50} & \textbf{99.65} \\
        \bottomrule
      \end{tabular}%
    }
  \end{subtable}
  
  \vspace{10pt}

  \begin{subtable}{\textwidth}
    \centering
    \caption{Detailed results of CVS robustness to target model (Qwen2-VL-2B) changes using a 10\% data subset.}
    \label{tab:tmsubset_10}
    \resizebox{\textwidth}{!}{%
      \begin{tabular}{lcccccccccccccc}
      \toprule
      \textbf{Method} & \textbf{GQA} & \textbf{VizWiz} & \textbf{TextVQA} & \textbf{SQA-I} & \textbf{MME} & \textbf{MMB-CN} & \textbf{MMB-EN} & \textbf{AI2D} & \textbf{ChartQA} & \textbf{DocVQA} & \textbf{InfoVQA} & \textbf{MMStar} & \textbf{OCRBench} & \textbf{ARP} \\
      \midrule
        Full Data & 51.84 & 34.21 & 72.03 & 74.96 & 1294.90 & 61.68 & 68.99 & 68.94 & 52.00 & 81.86 & 50.67 & 42.58 & 74.80 & 100.00 \\
        \midrule
        Random & \textbf{52.09} & 35.91 & 72.60 & 75.16 & \textbf{1398.61} & 55.58 & 62.03 & 67.88 & \textbf{50.96} & 78.10 & 51.61 & 40.36 & 74.30 & 98.64 \\
        \rowcolor{lightcyan}
        CVS & 48.37 & \textbf{44.26} & \textbf{73.57} & \textbf{75.36} & 1241.04 & \textbf{57.47} & \textbf{62.89} & \textbf{68.07} & 50.48 & \textbf{80.85} & \textbf{53.02} & \textbf{40.91} & \textbf{75.10} & \textbf{100.09} \\
        \bottomrule
      \end{tabular}%
    }
  \end{subtable}

  \vspace{10pt} 

  \begin{subtable}{\textwidth}
    \centering
    \caption{Detailed results of CVS robustness to target model (Qwen2-VL-2B) changes using a 15\% data subset.}
    \label{tab:tmsubset_15}
    \resizebox{\textwidth}{!}{%
      \begin{tabular}{lcccccccccccccc}
      \toprule
      \textbf{Method} & \textbf{GQA} & \textbf{VizWiz} & \textbf{TextVQA} & \textbf{SQA-I} & \textbf{MME} & \textbf{MMB-CN} & \textbf{MMB-EN} & \textbf{AI2D} & \textbf{ChartQA} & \textbf{DocVQA} & \textbf{InfoVQA} & \textbf{MMStar} & \textbf{OCRBench} & \textbf{ARP} \\
      \midrule
        Full Data & 51.84 & 34.21 & 72.03 & 74.96 & 1294.90 & 61.68 & 68.99 & 68.94 & 52.00 & 81.86 & 50.67 & 42.58 & 74.80 & 100.00 \\
        \midrule
        Random & 49.86 & \textbf{39.80} & 71.77 & 74.71 & 1194.15 & \textbf{60.30} & \textbf{66.15} & \textbf{68.55} & 44.44 & 77.15 & 51.18 & 40.06 & 74.40 & 97.80 \\
        \rowcolor{lightcyan}
        CVS & \textbf{52.08} & 36.21 & \textbf{74.27} & \textbf{75.16} & \textbf{1343.07} & 57.22 & 62.72 & 68.30 & \textbf{53.04} & \textbf{80.69} & \textbf{51.51} & \textbf{41.30} & \textbf{74.90} & \textbf{99.65} \\
        \bottomrule
      \end{tabular}%
    }
  \end{subtable}

\end{table*}

\begin{table*}[htbp]
  \centering
  \caption{Detailed results on Vision-Flan for robustness to the target model (LLaVA-1.5-13B). With the evaluator fixed to Qwen2-VL-7B-Instruct, we replace the target model from LLaVA-1.5-7B to LLaVA-1.5-13B to evaluate the robustness of CVS under target model changes. Experiments are conducted using data subsets of 5\%, 10\%, and 15\%. The best result in each column is highlighted in bold.}
  \label{tab:tm_13_full_comparison}

    \begin{subtable}{\textwidth}
      \centering
      \caption{Detailed results of CVS robustness to target model (LLaVA-1.5-13B) changes using a 5\% data subset.}
      \label{tab:tm13subset_5}
      \resizebox{\textwidth}{!}{%
        \begin{tabular}{lcccccccccccccc}
        \toprule
        \textbf{Method} & \textbf{GQA} & \textbf{VizWiz} & \textbf{TextVQA} & \textbf{SQA-I} & \textbf{MME} & \textbf{MMB-CN} & \textbf{MMB-EN} & \textbf{AI2D} & \textbf{ChartQA} & \textbf{DocVQA} & \textbf{InfoVQA} & \textbf{MMStar} & \textbf{OCRBench} & \textbf{ARP} \\
        \midrule
        Full Data 
        & 49.02 & 55.92 & 35.82 & 70.60 & 1318.89 & 56.44 & 60.74 & 57.67 & 19.44 & 20.03 & 19.87 & 35.65 & 29.50 & 100.00 \\
        \midrule
        Random 
        & \textbf{43.02} & \textbf{56.82} & 34.35 & \textbf{63.46} & \textbf{1178.91} & 36.25 & 41.23 & \textbf{45.66} & 14.88 & 14.87 & 18.17 & 32.17 & \textbf{29.00} & 85.12 \\
        \rowcolor{lightcyan}
        CVS 
        & 39.54 & 56.80 & \textbf{36.64} & 63.21 & 920.15 & \textbf{39.35} & \textbf{45.37} & 43.65 & \textbf{18.12} & \textbf{18.24} & \textbf{21.95} & \textbf{33.50} & 28.40 & \textbf{88.34} \\
        \bottomrule
        \end{tabular}%
      }
    \end{subtable}
  
  \vspace{10pt}

    \begin{subtable}{\textwidth}
      \centering
      \caption{Detailed results of CVS robustness to target model (LLaVA-1.5-13B) changes using a 10\% data subset.}
      \label{tab:tms13ubset_10}
      \resizebox{\textwidth}{!}{%
        \begin{tabular}{lcccccccccccccc}
        \toprule
        \textbf{Method} & \textbf{GQA} & \textbf{VizWiz} & \textbf{TextVQA} & \textbf{SQA-I} & \textbf{MME} & \textbf{MMB-CN} & \textbf{MMB-EN} & \textbf{AI2D} & \textbf{ChartQA} & \textbf{DocVQA} & \textbf{InfoVQA} & \textbf{MMStar} & \textbf{OCRBench} & \textbf{ARP} \\
        \midrule
        Full Data & 49.02 & 55.92 & 35.82 & 70.60 & 1318.89 & 56.44 & 60.74 & 57.67 & 19.44 & 20.03 & 19.87 & 35.65 & 29.50 & 100.00 \\
        \midrule
        Random &
        \textbf{44.98} & \textbf{58.13} & 36.43 & \textbf{68.26} & \textbf{1176.93} & \textbf{47.60} & \textbf{55.56} & \textbf{49.63} & 16.56 & 16.97 & 19.08 & 31.35 & 29.60 & 92.26 \\
        \rowcolor{lightcyan}
        CVS &
        41.27 & 57.48 & \textbf{37.49} & 66.58 & 1038.90 & 43.04 & 48.45 & 47.80 & \textbf{17.60} & \textbf{19.03} & \textbf{21.73} & \textbf{35.66} & \textbf{29.80} & \textbf{92.28} \\
        \bottomrule
        \end{tabular}%
      }
    \end{subtable}

  \vspace{10pt} 

    \begin{subtable}{\textwidth}
      \centering
      \caption{Detailed results of CVS robustness to target model (LLaVA-1.5-13B) changes using a 15\% data subset.}
      \label{tab:tm13subset_15}
      \resizebox{\textwidth}{!}{%
        \begin{tabular}{lcccccccccccccc}
        \toprule
        \textbf{Method} & \textbf{GQA} & \textbf{VizWiz} & \textbf{TextVQA} & \textbf{SQA-I} & \textbf{MME} & \textbf{MMB-CN} & \textbf{MMB-EN} & \textbf{AI2D} & \textbf{ChartQA} & \textbf{DocVQA} & \textbf{InfoVQA} & \textbf{MMStar} & \textbf{OCRBench} & \textbf{ARP} \\
        \midrule
        Full Data & 49.02 & 55.92 & 35.82 & 70.60 & 1318.89 & 56.44 & 60.74 & 57.67 & 19.44 & 20.03 & 19.87 & 35.65 & 29.50 & 100.00 \\
        \midrule
        Random & \textbf{45.25} & \textbf{56.96} & \textbf{38.03} & 65.05 & \textbf{1176.15} & 42.70 & 46.99 & \textbf{48.32} & \textbf{16.60} & 18.38 & 18.23 & \textbf{34.86} & 29.40 & 91.14 \\
        \rowcolor{lightcyan}
        CVS & 42.54 & 56.55 & 37.02 & \textbf{67.18} & 1150.28 & \textbf{44.07} & \textbf{51.63} & 46.11 & 15.80 & \textbf{19.07} & \textbf{21.24} & 32.30 & \textbf{30.50} & \textbf{91.85} \\
        \bottomrule
        \end{tabular}%
      }
    \end{subtable}

\end{table*}

\begin{table*}[htbp]
  \centering
    \caption{Detailed results on Vision-Flan for visual anchoring.
    To evaluate the necessity of the visual anchoring term in CVS, we remove the image information $I$ from the conditional probability $P(Y \mid I, A)$ in the denominator and compare the resulting variant with random sampling using data subsets of 5\%, 10\%, and 15\%. The best result in each column is highlighted in bold.}
    
  \label{tab:va_full_comparison}

  \begin{subtable}{\textwidth}
    \centering
    \caption{Detailed results of the visual anchoring ablation study using a 5\% data subset.}
    \label{tab:subset_5}
    \resizebox{\textwidth}{!}{%
      \begin{tabular}{lcccccccccccccc}
      \toprule
      \textbf{Method} & \textbf{GQA} & \textbf{VizWiz} & \textbf{TextVQA} & \textbf{SQA-I} & \textbf{MME} & \textbf{MMB-CN} & \textbf{MMB-EN} & \textbf{AI2D} & \textbf{ChartQA} & \textbf{DocVQA} & \textbf{InfoVQA} & \textbf{MMStar} & \textbf{OCRBench} & \textbf{ARP} \\
      \midrule
        Full Data & 47.30 & 55.88 & 35.83 & 61.43 & 1270.40 & 50.26 & 55.67 & 52.14 & 15.68 & 15.80 & 15.53 & 35.32 & 26.20 & 100.00 \\
        \midrule
        Random & \textbf{42.93} & 55.79 & \textbf{36.54} & \textbf{60.34} & \textbf{1098.02} & \textbf{36.40} & \textbf{26.19} & 37.82 & \textbf{15.96} & \textbf{16.46} & \textbf{19.99} & \textbf{34.77} & 27.80 & \textbf{92.96} \\
        \rowcolor{lightcyan}
        CVS & 36.12 & \textbf{56.29} & 33.08 & 57.56 & 493.39 & 31.19 & 14.69 & \textbf{37.89} & 14.92 & 15.99 & 17.29 & 33.08 & \textbf{28.10} & 82.44 \\
        \bottomrule
      \end{tabular}%
    }
  \end{subtable}
  
  \vspace{10pt}

  \begin{subtable}{\textwidth}
    \centering
    \caption{Detailed results of the visual anchoring ablation study using a 10\% data subset.}
    \label{tab:subset_10}
    \resizebox{\textwidth}{!}{%
      \begin{tabular}{lcccccccccccccc}
      \toprule
      \textbf{Method} & \textbf{GQA} & \textbf{VizWiz} & \textbf{TextVQA} & \textbf{SQA-I} & \textbf{MME} & \textbf{MMB-CN} & \textbf{MMB-EN} & \textbf{AI2D} & \textbf{ChartQA} & \textbf{DocVQA} & \textbf{InfoVQA} & \textbf{MMStar} & \textbf{OCRBench} & \textbf{ARP} \\
      \midrule
        Full Data & 47.30 & 55.88 & 35.83 & 61.43 & 1270.40 & 50.26 & 55.67 & 52.14 & 15.68 & 15.80 & 15.53 & 35.32 & 26.20 & 100.00 \\
        \midrule
        Random & \textbf{43.31} & \textbf{56.02} & \textbf{36.13} & 60.92 & \textbf{1152.77} & \textbf{39.61} & 31.88 & \textbf{44.50} & \textbf{16.52} & 17.27 & 18.08 & \textbf{34.36} & 27.90 & \textbf{95.30} \\
        \rowcolor{lightcyan}
        CVS & 35.03 & 52.96 & 33.35 & \textbf{64.06} & 501.64 & 36.17 & \textbf{41.75} & 41.22 & 14.48 & \textbf{18.15} & \textbf{19.80} & 30.84 & \textbf{28.70} & 89.48 \\
        \bottomrule
      \end{tabular}%
    }
  \end{subtable}

  \vspace{10pt} 

  \begin{subtable}{\textwidth}
    \centering
    \caption{Detailed results of the visual anchoring ablation study using a 15\% data subset.}
    \label{tab:subset_15}
    \resizebox{\textwidth}{!}{%
      \begin{tabular}{lcccccccccccccc}
      \toprule
      \textbf{Method} & \textbf{GQA} & \textbf{VizWiz} & \textbf{TextVQA} & \textbf{SQA-I} & \textbf{MME} & \textbf{MMB-CN} & \textbf{MMB-EN} & \textbf{AI2D} & \textbf{ChartQA} & \textbf{DocVQA} & \textbf{InfoVQA} & \textbf{MMStar} & \textbf{OCRBench} & \textbf{ARP} \\
        \midrule
        Full Data & 47.30 & 55.88 & 35.83 & 61.43 & 1270.40 & 50.26 & 55.67 & 52.14 & 15.68 & 15.80 & 15.53 & 35.32 & 26.20 & 100.00 \\
        \midrule
        Random & \textbf{43.19} & \textbf{56.33} & \textbf{35.41} & 59.59 & \textbf{1230.04} & \textbf{42.44} & 38.06 & 41.06 & \textbf{15.52} & 16.26 & 16.26 & \textbf{32.27} & \textbf{27.60} & \textbf{93.82} \\
        \rowcolor{lightcyan}
        CVS & 35.02 & 53.42 & 30.81 & \textbf{63.31} & 499.41 & 41.15 & \textbf{46.82} & \textbf{41.97} & 14.28 & \textbf{17.58} & \textbf{17.04} & 31.42 & 27.30 & 88.44 \\
        \bottomrule
      \end{tabular}%
    }
  \end{subtable}

\end{table*}

\clearpage

\vspace{-6pt}
\clearpage

\end{document}